\documentclass{article} 
\usepackage{iclr2027_conference,times}

\usepackage{amsmath,amsfonts,bm}

\def\eqref#1{equation~\ref{#1}}

\def\1{\bm{1}}

\DeclareMathAlphabet{\mathsfit}{\encodingdefault}{\sfdefault}{m}{sl}
\SetMathAlphabet{\mathsfit}{bold}{\encodingdefault}{\sfdefault}{bx}{n}

\usepackage{hyperref}
\usepackage{url}
\usepackage{graphicx}
\usepackage{booktabs}
\usepackage{multirow}
\usepackage{bbding}
\usepackage{pifont}
\usepackage{ulem} 
\usepackage{wrapfig}
\usepackage{graphicx}
\usepackage{fontawesome5}
\newcommand{\corresponding}{%
  \raisebox{0.15ex}{\scriptsize\faEnvelope}%
}

\title{DailyBench: A Unified Benchmark for AI-Generated and Manipulated Images from Modern Generative Models}

\author{
Xin Jiang$^{1}$,
Hao Tang$^{2,\dagger}$,
Junyao Gao$^{3}$,
Zijie Yang$^{1}$,
Meiqi Cao$^{4}$\\
\textbf{
Fei Shen$^{4}$,
Dongming Zhang$^{1}$,
Zechao Li$^{4}$,
Yongdong Zhang$^{5,\corresponding}$
}\\[2mm]
$^{1}$State Key Laboratory of Communication Content Cognition, People's Daily Online \\
$^{2}$Nanyang Technological University,
$^{3}$Tongji University \\
$^{4}$Nanjing University of Science and Technology,
$^{5}$University of Science and Technology of China \\
$^{\dagger}$Project Leader \qquad
$^{\corresponding}$Corresponding Author \\
\texttt{jiangxin@people.cn, howard.haotang@gmail.com, zhyd73@ustc.edu.cn}
}

\usepackage[table]{xcolor}  
\definecolor{mygray}{gray}{.9}

\begin{document}

\maketitle
\begin{figure*}[htbp]
\vspace{-8mm}
    \centering
\includegraphics[width=1.0\linewidth]{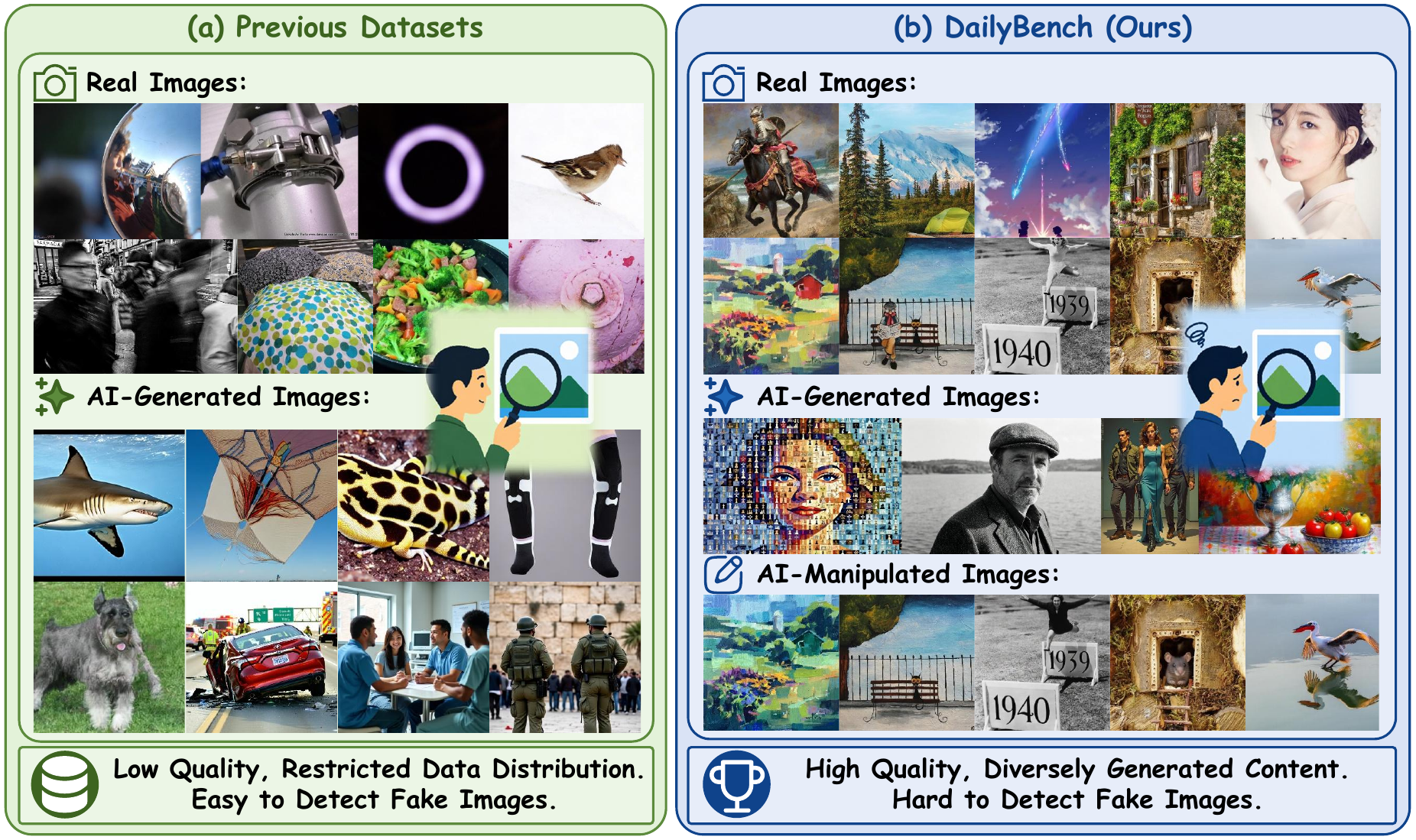}
\vspace{-4mm}
    \caption{Comparison between \textbf{DailyBench} and previous AIID benchmarks. Existing benchmarks such as GenImage~\cite{genimage} and RRDataset~\cite{rrdataset} are constructed with outdated generators and restricted data distributions, whereas \textbf{DailyBench} incorporates modern generators such as Qwen-Image~\cite{qwenimage} and Nano Banana 2~\cite{NanoBanana} to cover more diverse synthetic and manipulated images.}
    \label{fig:dataset_compare}
    \vspace{-2mm}
\end{figure*}
\begin{abstract}
    Recent advances in generative models have shifted AI-generated image detection from identifying easily distinguishable, fully synthetic images to identifying highly realistic content generated by both modern generation and manipulation pipelines. 
    However, existing detection benchmarks are often built with outdated generative models and primarily emphasize full-image synthesis, creating a growing mismatch between benchmark data and the images encountered in real-world generation and editing scenarios.
    To bridge this gap, we introduce \textbf{DailyBench}, a high-quality unified benchmark for evaluating whether AI-generated image detectors can generalize across both modern full-image synthesis and object-level manipulation.
    DailyBench contains two complementary subsets: \textbf{FakeBench}, which includes high-quality images synthesized by recent open-source and commercial generative models, and \textbf{ManipulationBench}, which introduces challenging object-level edits applied to real images using advanced image-conditional models.
    This design makes DailyBench a realistic testbed for studying both generator-level generalization and manipulation-aware detection under subtle local edits.
    In addition to the benchmark, we propose \textbf{FPD}~(\textbf{F}ake-\textbf{P}reference \textbf{D}etector), a diagnostic-driven strong baseline that encourages diagnostic-sensitive representation learning.
    FPD combines a dual-pathway architecture with a cascaded compression classifier to extract and compact diagnostic-relevant cues progressively, and further introduces a fake-preference sampler to reduce excessive bias toward the real class.
    Experiments on DailyBench reveal substantial robustness gaps in current detectors: methods reporting 91-96\% balanced accuracy on GenImage drop to 52-79\% on FakeBench and 43-67\% on ManipulationBench.
    These results show that existing detectors remain poorly generalized to realistic synthesis and manipulation, highlighting DailyBench as a rigorous testbed for developing robust and manipulation-aware AI-generated image detection methods.
    %
    %
\end{abstract}

\section{Introduction}
The rapid development of image generative models~\citep{sd,flux2024,qwenimage,NanoBanana} has made synthetic visual content increasingly realistic and accessible. Images generated or edited by modern generative models are often difficult to distinguish from real photographs, enabling valuable applications in creative production and visual design while also introducing serious risks, including misinformation, digital fraud, and malicious content generation. Therefore, AI-generated image detection (AIID), which aims to distinguish real images from AI-generated or AI-manipulated images, has become an important research direction for cognitive security in content communication.

While most existing AIID benchmarks provide a valuable foundation, but they suffer from several important limitations.
(1) \textbf{Limited diversity and quality of real images.} Many benchmarks rely on restricted real-image sources, which limits their ability to measure detector generalization across diverse domains and real-world scenes. For example, the real images in GenImage~\cite{genimage} are collected from ImageNet~\cite{imagenet}, whereas B-Free~\cite{bias-free} is constructed from COCO~\cite{coco}; both sources provide useful annotations but cover limited visual distributions for open-domain AIID evaluation.
(2) \textbf{Outdated generative models.} The rapid advancement of generative models has made many existing benchmarks increasingly outdated, as shown in Tab.~\ref{tab:dataset_diff}. Consequently, their generated images often have lower visual quality, as illustrated in Fig.~\ref{fig:dataset_compare}, and cannot accurately reflect the detection difficulty introduced by modern generative models~\cite{NanoBanana}.
(3) \textbf{Limited coverage of AI-generated content.} Existing datasets primarily focus on fully synthesized images and often overlook more subtle and realistic forms of manipulation, such as local edits and object-level modifications. These localized manipulations are particularly challenging because they preserve most real-image content while altering only specific regions, making them more likely to evade detectors and deceive users in real-world scenarios.
%
%
These limitations motivate a benchmark that combines diverse high-quality real images, modern generative models, and coverage of both full-image synthesis and localized manipulation.

\begin{table}[htbp]
    \centering
    \vspace{-6mm}
    \caption{Comparison of \textbf{DailyBench} with existing AI-generated image detection benchmarks. \textbf{AS}, \textbf{MR}, and \textbf{AR} denote average aesthetic score, minimum resolution, and average resolution, respectively. \textbf{Mani.I} denotes manipulated images, and \textbf{Gen.T} denotes the release year of the latest generator included in each benchmark.}
    \resizebox{1.0\textwidth}{!}{
    \begin{tabular}{lccccccccc}
    \toprule
    \multirow{2}{*}{Dataset}& \multicolumn{3}{c}{Real Image} & \multicolumn{3}{c}{Fake Image} &\multirow{2}{*}{Mani.I} &\multirow{2}{*}{Gen.T}&\multirow{2}{*}{Number} 
    \\\cmidrule(lr){2-4}\cmidrule(lr){5-7}
             &AS  &MR &AR      &AS  &MR &AR
             & & &
             \\
             \midrule
         GenImage~\cite{genimage} &5.43  &54 & 490$\times$430      &4.90 &128 &435$\times$435  
             & \ding{55} &2022 &100K\\
         AIGCDetect~\cite{AIGCDetect} &5.50  &54 &427$\times$382      &5.03  &256 &427$\times$420
             &\ding{55} &2022 &152K \\
         WildRF~\cite{wildrf} &5.83  &144 &2171$\times$ 2208      &6.45  &255 &1284$\times$1261
             &\ding{55} &Unknown &5.3K \\
             SynthWildX~\cite{sysnxwidx} &5.99  &231 &1505$\times$1459      &6.68  &256 &1344$\times$1236
             & \ding{55}  &2023 &2K \\
         Chameleon~\cite{chameleon} &5.76  &92 &500$\times$496      &6.39  &128 &699$\times$901
             &\ding{55} &Unknown &26K \\
         RRDataset~\cite{rrdataset} &5.46  &141 &1345$\times$1101      &5.85  &151 &660$\times$681
             & \ding{55} &2024 &12K \\
         AIGIBench~\cite{aigibench}&5.51  &60   &439$\times$417     &4.92  &256 &409$\times$409
             &\ding{55} &2024 &26K \\
         
         \midrule
         \textbf{DailyBench~(Ours)}  &6.87  &512 &512$\times$512      &7.01  &512 &777$\times$774
             &\ding{51} &2026 &270K\\
          \bottomrule
    \end{tabular}
    }
    \label{tab:dataset_diff}
\end{table}

To address these limitations, we introduce \textbf{DailyBench}, a high-quality unified benchmark suite containing 140K AI-generated and AI-manipulated images generated by modern generative models.
Rather than merely increasing dataset scale, DailyBench is designed around three properties required for realistic AIID evaluation: \textbf{visual quality}, \textbf{modern generator coverage}, and \textbf{generated-content diversity}.
First, DailyBench improves visual quality by filtering both real and AI-generated images with an aesthetic predictor~\cite{schuhmann2022laion}, removing low-quality samples with low aesthetic scores.
Second, DailyBench expands generator coverage by incorporating five modern text-to-image models and six modern image-conditional generation models, including both open-source and commercial closed-source systems.
Third, DailyBench broadens generated-content diversity through two complementary subsets: \textbf{FakeBench}, which consists of fully AI-generated images synthesized from text prompts, and \textbf{ManipulationBench}, which consists of AI-manipulated images generated from real images with text-guided editing prompts.
Together, these designs enable DailyBench to evaluate AIID methods in both full-image synthesis and localized-manipulation settings.
A detailed comparison between DailyBench and existing AIID benchmarks is shown in Tab.~\ref{tab:dataset_diff}.

Beyond DailyBench, we further propose \textbf{F}ake-\textbf{P}reference \textbf{D}etector \textbf{(FPD)}, a diagnostic-driven strong baseline for AIID.
FPD uses a dual-pathway architecture to capture diagnostic cues from both semantic and latent image representations.
The extracted features are then fed into a cascaded compression classifier, which progressively condenses diagnostic-related information for final prediction.
During training, a fake-preference sampler places more fake than real images in each mini-batch, increasing the fake class's contribution to the loss to address the observed preference for real-image predictions.
%

Our experiments on DailyBench reveal a clear gap between existing benchmark performance and robustness under modern generation and manipulation settings.
Although several detectors achieve 91-96\% balanced accuracy on GenImage, their performance drops to only 52-79\% on FakeBench and 43-67\% on ManipulationBench.
This degradation indicates that many existing detectors fail to generalize from earlier benchmark distributions to realistic fake images and localized object-level manipulations.
Training with data from modern generators~\cite{qwenimage} improves performance on FakeBench for many methods.
These findings demonstrate the necessity of DailyBench as a more rigorous testbed and highlight the need for AIID methods that are both robust across generation sources and sensitive to localized manipulations.

In summary, our contributions are as follows:
\begin{enumerate}
    \item We introduce \textbf{DailyBench}, a high-quality unified benchmark suite for comprehensive AIID evaluation, containing 140K AI-generated and AI-manipulated images generated by modern generative models.
    \item We construct two complementary subsets, \textbf{FakeBench} and \textbf{ManipulationBench}, covering both fully synthetic images and realistic localized manipulations to evaluate detector robustness under modern generation and editing scenarios.
     \item We propose \textbf{F}ake-\textbf{P}reference \textbf{D}etector \textbf{(FPD)}, a strong diagnostic-driven baseline that combines dual-pathway representation learning, a cascaded compression classifier, and a fake-preference sampler for AIID.
    \item We conduct extensive experiments showing that existing AIID methods suffer substantial performance degradation on DailyBench, revealing a significant gap between existing benchmark performance and real-world generalization.
   
\end{enumerate}
\section{DailyBench} \label{sec_daily}
\begin{figure*}
    \centering
    \includegraphics[width=0.9\linewidth]{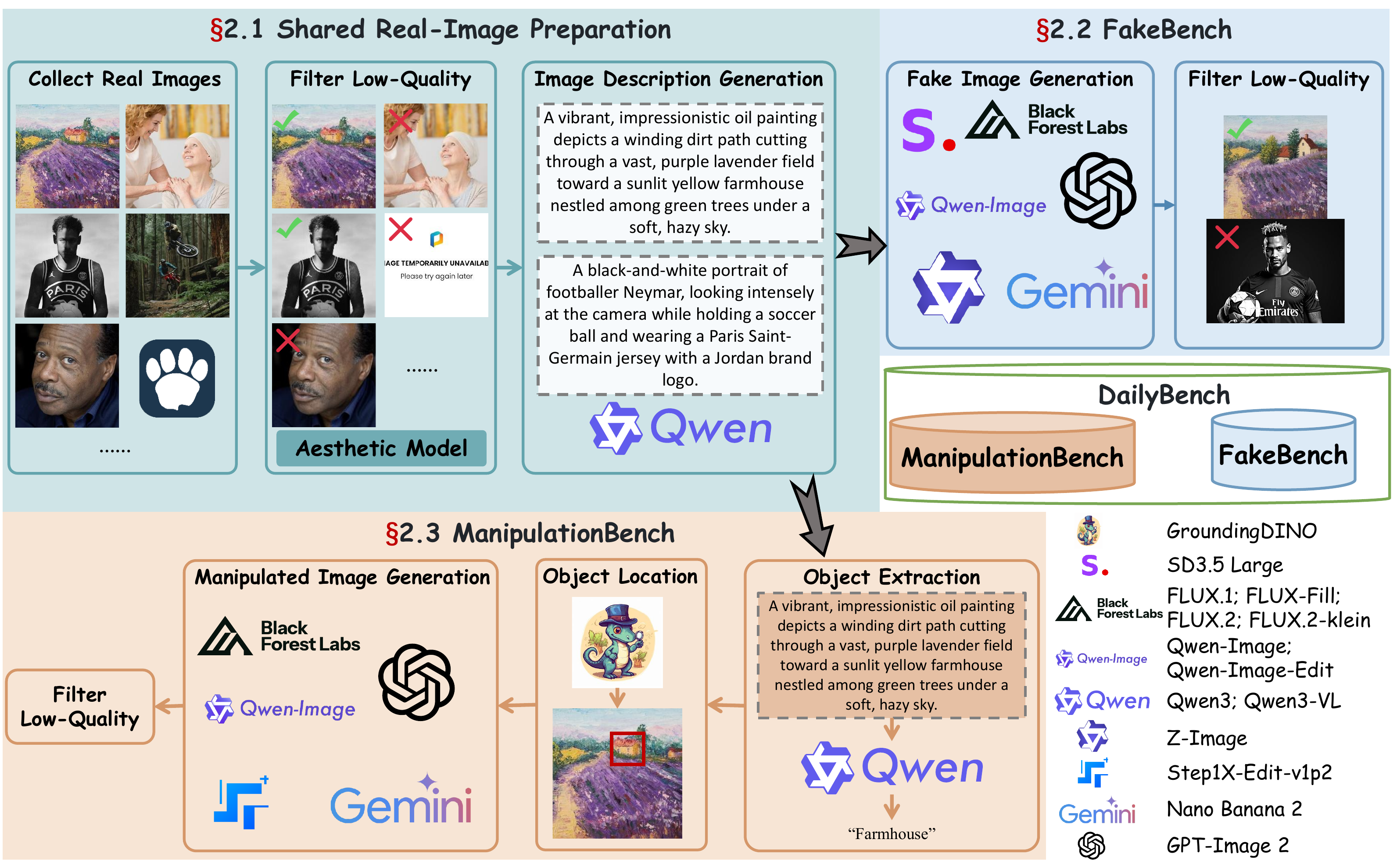}
     \vspace{-2mm}
    \caption{Overview of DailyBench. For more details, please refer to Sec.~\ref{sec_daily}.}
    \label{fig:pipeline}
     \vspace{-6mm}
\end{figure*}

Rather than merely increasing dataset scale, \textbf{DailyBench} is designed around three benchmark construction principles: high visual quality, up-to-date generator coverage, and comprehensive coverage of synthetic content types.
To improve visual quality, we filter all images so that each image has a minimum resolution of 512 pixels on the shortest side and an aesthetic score of at least 6.
To reflect the capabilities of modern generative models, DailyBench covers both recent text-to-image~(T2I) models and image-conditional editing models. Specifically, DailyBench includes five recent open-source T2I models, including SD3.5 Large~\cite{sd3.5}, FLUX.1~\cite{flux2024}, FLUX.2~\cite{flux2}, Qwen-Image-2512~\cite{qwenimage}, and Z-Image~\cite{z-image}, together with two proprietary models, Nano Banana 2~\cite{NanoBanana} and GPT-Image 2~\cite{gptimage2}; four recent open-source image-conditional models, including FLUX-Fill~\cite{flux2024}, FLUX.2-klein-9B~\cite{flux2}, Qwen-Image-Edit-2511~\cite{qwenimage}, and Step1X-Edit-v1p2~\cite{stepedit}.
To cover different forms of AI-generated content, DailyBench contains two complementary subsets: \textbf{FakeBench}, which targets fully synthetic images, and \textbf{ManipulationBench}, which targets localized image manipulations.
Together, these subsets allow DailyBench to evaluate detector robustness across both global generation artifacts and subtle object-level editing artifacts.
The overall construction pipeline is shown in Fig.~\ref{fig:pipeline}. We next describe the shared real-image preparation process, followed by the construction of FakeBench and ManipulationBench.

\subsection{Shared Real-Image Preparation}
A key requirement for building a reliable AI-generated image detection benchmark is a diverse and high-quality collection of real-world images.
Many existing benchmarks~\cite{genimage,bias-free,chen2024drct} primarily rely on ImageNet~\cite{imagenet} or COCO~\cite{coco}, which restricts the diversity of source content and consequently limits the diversity of synthesized images.
To obtain a broader real-image foundation, we curate images from LAION-Aesthetics V2~\cite{schuhmann2022laion}, a large-scale image collection with aesthetic quality annotations.
We preserve images with an aesthetic score of at least 6 and a shortest-side resolution of at least 512 pixels.
This filtering process improves the visual quality of the real-image pool and provides high-quality source images for both full-image synthesis and localized manipulation.
For each preserved real image, we use Qwen3-VL-8B~\cite{bai2025qwen3vl} to generate a textual description $\mathbf{T}$.
These descriptions are used as prompts for generating fake images in FakeBench and as semantic guidance for constructing manipulated images in ManipulationBench.

\subsection{FakeBench}
FakeBench targets the full-image synthesis setting and is designed to evaluate whether AI-generated image detectors can generalize across modern T2I generators.
Given the prepared real-image collection, we construct synthetic counterparts using seven recent T2I models: five open-source models, including SD3.5 Large~\cite{sd3.5}, FLUX.1~\cite{flux2024}, FLUX.2~\cite{flux2}, Qwen-Image-2512~\cite{qwenimage}, and Z-Image~\cite{z-image}, and two proprietary models, Nano Banana 2~\cite{NanoBanana} and GPT-Image 2~\cite{gptimage2}.
These models differ in architecture, parameter scale, training distribution, and generation behavior, which introduces diverse visual styles and generator-specific artifacts.
By covering this generator diversity, FakeBench provides a challenging testbed for measuring detector robustness to modern full-image synthesis rather than overfitting to artifacts from a single generator family.

\begin{figure*}
    \centering
    \includegraphics[width=0.9\linewidth]{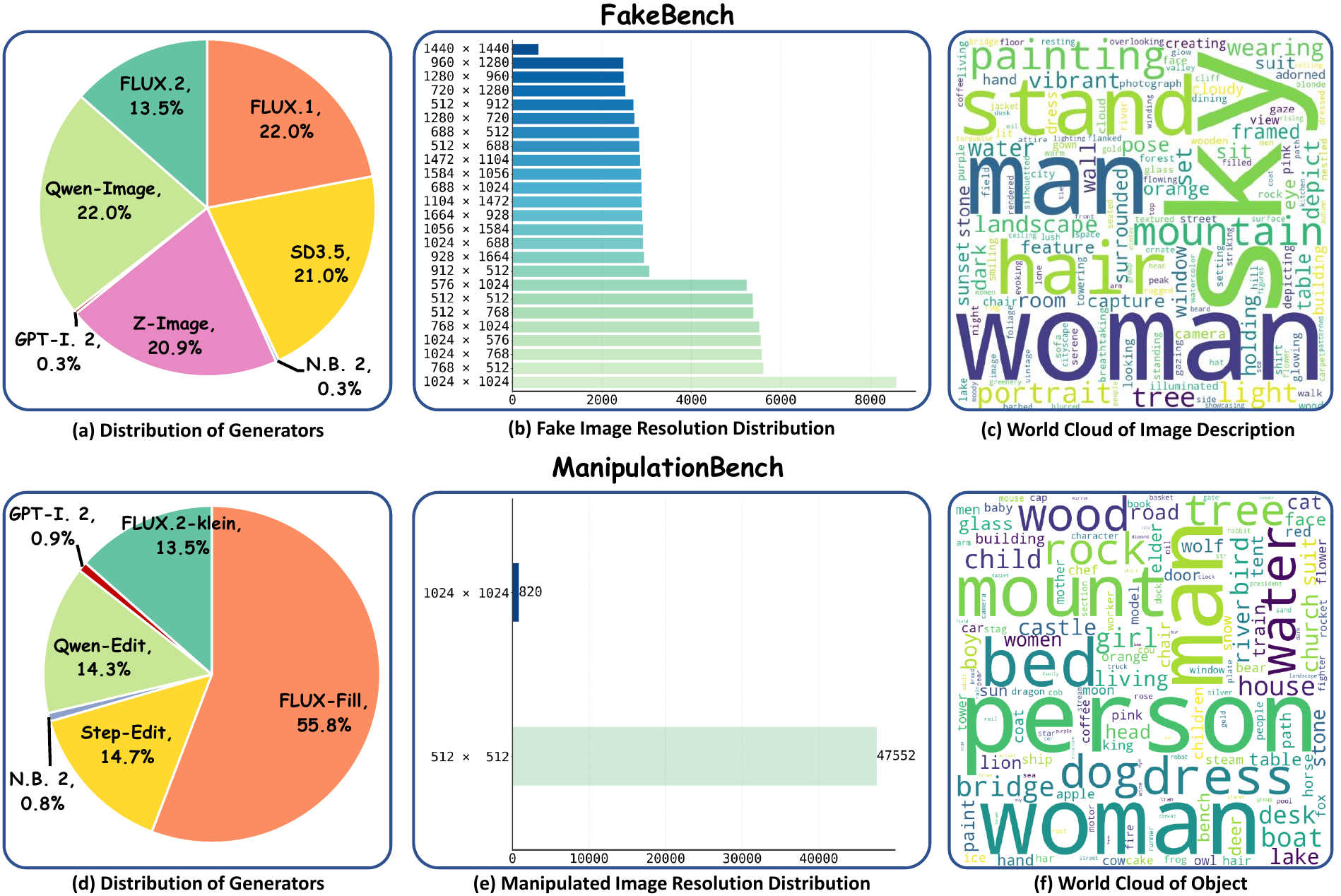}
    \caption{Statistics of the FakeBench and ManipulationBench. (a) Distribution of samples generated by different generators in the FakeBench. (b) Distribution of fake image resolution in the FakeBench. (c) World cloud of the image description in the FakeBench. (d) Distribution of samples generated by different generators in the ManipulationBench. (e) Distribution of the manipulated image resolution in the ManipulationBench. (f) World cloud of the object in the ManipulationBench.}
    \label{fig:datadistribution}
     \vspace{-4mm}
\end{figure*}

\textbf{Fake Image Generation.}
Given the caption $\mathbf{T}$ of a real image, we generate a corresponding fake image using a T2I generator:
\begin{equation}
    \mathbf{F} = \mathrm{G}(\mathbf{T}),
\end{equation}
where $\mathrm{G}$ denotes a T2I generative model.
To match the output characteristics of modern generators, we keep the shortest-side resolution of each synthesized image no smaller than 512 pixels.
We further adopt a diversified resolution setting, as shown in Fig.~\ref{fig:datadistribution}(b), with resolutions ranging from 0.5K to 1K and common aspect ratios including 1:1, 3:4, 9:16, and 3:2.

\textbf{Quality Filtering.}
Although recent generators can generate highly realistic images, they may still create visually implausible or low-quality samples.
To reduce such confounding cases, we apply the Aesthetic Predictor~\cite{schuhmann2022laion} and discard synthesized images with aesthetic scores below 6.
This filtering step improves the overall visual quality of FakeBench and ensures that detectors are evaluated on realistic fake images rather than trivially flawed generations.

\subsection{ManipulationBench}
ManipulationBench is designed to evaluate AI-generated image detection under localized image manipulation, where only selected regions are synthesized while most real-image content is preserved.
This setting has become increasingly practical with recent image-conditional generative models~\cite{flux2,qwenimage,stepedit}, but it remains challenging for detectors because manipulated images contain partial synthetic content and often exhibit fewer global artifacts than fully synthesized images.
To cover diverse editing mechanisms, ManipulationBench includes AI-manipulated images generated by both inpainting-based and editing-based image-conditional models.
We use four recent open-source models, including FLUX-Fill~\cite{flux2024}, FLUX.2-klein-9B~\cite{flux2}, Qwen-Image-Edit-2511~\cite{qwenimage}, and Step1X-Edit-v1p2~\cite{stepedit}, together with two proprietary systems, Nano Banana 2~\cite{NanoBanana} and GPT-Image 2~\cite{gptimage2}.
These models cover different manipulation paradigms and output distributions, enabling ManipulationBench to evaluate detector robustness under realistic localized editing scenarios.
%

\textbf{Inpainting-based Manipulation.}
For the inpainting-based image-conditional model FLUX-Fill, we construct two manipulation settings using random masks and object masks.
The random mask $\mathbf{M}_\textrm{R}$ is a rectangular region covering 30\% of the image area, while the object mask $\mathbf{M}_\textrm{O}$ is derived from the bounding box of a localized object.
For random-mask inpainting, we use the image description $\mathbf{T}$ as the conditional prompt, encouraging the model to fill the masked region consistently with the original image context.
For object-level manipulation, we consider two operations: object removal and object replacement.
For object removal, we use the prompt $\mathbf{T}_\textrm{M}$: \textit{``Only extend the existing background into the masked region. Do not generate new objects, people, text, patterns, or structures.''}
For object replacement, we use the prompt $\mathbf{T}_\textrm{S}$: \textit{``Object''}, where \textit{Object} denotes the target object category to be synthesized.
The generation process is formulated as:
\begin{equation}
\mathbf{F}_\textrm{R} = \mathrm{FLUX\text{-}Fill}(\mathbf{T}, \mathbf{I}, \mathbf{M}_\textrm{R}), \quad
\mathbf{F}_\textrm{M/S} = \mathrm{FLUX\text{-}Fill}(\mathbf{T}_\textrm{M/S}, \mathbf{I}, \mathbf{M}_\textrm{O}),
\end{equation}
where $\mathbf{I}$ denotes the input real image, $\mathbf{F}_\textrm{R}$ denotes the random-mask inpainting output, and $\mathbf{F}_\textrm{M}$ and $\mathbf{F}_\textrm{S}$ denote the object-removal and object-replacement outputs, respectively.

\textbf{Editing-based Manipulation.}
For editing-based image-conditional models, we use the input image and a text editing instruction as conditions.
We define two instruction types for each localized target object: object removal and object replacement.
For object removal, we use the instruction $\mathbf{P}_\textrm{M}$: \textit{``Remove the \{Object\}''}.
For object replacement, we use the instruction $\mathbf{P}_\textrm{S}$: \textit{``Replace the \{Object\} with a new \{Object\}''}.
Here, \textit{\{Object\}} denotes the localized target object selected from the input image.
The generation process is formulated as:
\begin{equation}
    \mathbf{F}_\textrm{M/S} = \textrm{IG}(\mathbf{P}_\textrm{M/S}, \mathbf{I}),
\end{equation}
where $\mathrm{IG}(\cdot, \cdot)$ denotes an image-editing generative model, $\mathbf{I}$ denotes the input real image, and $\mathbf{F}_\textrm{M}$ and $\mathbf{F}_\textrm{S}$ denote the object-removal and object-replacement outputs, respectively.
Following FakeBench, we apply aesthetic filtering and preserve samples with aesthetic scores above 6 to construct the ManipulationBench.

\subsection{Benchmark Statistics}
We report the statistics of DailyBench in Fig.~\ref{fig:datadistribution} to characterize its generator coverage, image quality, and content diversity.
Specifically, the generator distribution summarizes the recent T2I and image-conditional models used to construct DailyBench, reflecting its coverage of both generation and manipulation sources.
The resolution distribution verifies the high visual quality of the benchmark after image filtering, while the word-cloud visualization shows the diversity of image content and scene categories.
In addition, Tab.~\ref{tab:dataset_diff} compares DailyBench with prior AIID benchmarks, showing that DailyBench uses higher-quality real images, incorporates more recent generative models, and covers both fully synthetic images and localized manipulations.
These statistics and comparisons demonstrate that DailyBench provides a more comprehensive evaluation setting for AIID methods under modern generation and manipulation scenarios.
More details are provided in Supp.~\ref{bench_supp}.
\section{Diagnostic-Driven Strong Baseline}
Please see Supp.~\ref{sup_fpd}
\section{Experiments}
\subsection{Experiment Settings}
\textbf{Datasets.} 
We evaluate AIID methods under both standard and out-of-domain settings.
For standard evaluation, we use GenImage~\cite{genimage}, a widely adopted benchmark for AIID.
For in-the-wild evaluation, we use Chameleon~\cite{chameleon}, SynthWildX~\cite{sysnxwidx}, WildRF~\cite{wildrf}, and RRDataset~\cite{rrdataset}.
These datasets contain images collected from complex online environments and include realistic degradations caused by image compression, propagation, and re-sharing on the web.
To further study whether training data from modern generators improves robustness to modern AI-generated images, we construct \textbf{Qwen-Image-Train}, a training set containing 24K real images and fake images generated by Qwen-Image~\cite{qwenimage}.

\textbf{Baseline Methods.}
We evaluate a range of recent and representative AI-generated image detection methods on DailyBench, including UniFD~\cite{unifd}, DRCT~\cite{chen2024drct}, FatFormer~\cite{fatformer}, NPR~\cite{npr}, SAFE~\cite{safe}, Effort~\cite{effort}, Forgelens~\cite{chen2025forgelens}, AIDE~\cite{chameleon}, Aligned~\cite{rajan2025aligned}, B-Free~\cite{bias-free}, LOTA~\cite{wang2025lota}, DINO-Det~\cite{dinodet}, OmniAID~\cite{guo2025omniaid}, DDA~\cite{dda}, PROBE~\cite{probe}, and LTD~\cite{ltd}.

\subsection{Benchmark Results and Analysis}
\textbf{FakeBench.}
Tab.~\ref{tab:fake_compare} reports the performance of existing detectors on FakeBench.
Compared with their performance on GenImage~(Tab.~\ref{tab:genimage}), most detectors experience a substantial drop in accuracy, indicating that FakeBench contains more challenging images generated by modern T2I models.
Many methods also show a clear bias toward the real class, which supports the motivation for the fake-preference sampling strategy in FPD.
Early detectors such as UniFD~\cite{unifd}, FatFormer~\cite{fatformer}, and AIDE~\cite{chameleon} show limited discriminative ability on images generated by modern T2I models, while stronger recent detectors still exhibit weak performance.
These results suggest that training and evaluation data must better reflect the rapidly evolving distribution of modern generative models.
FPD achieves an average accuracy of 79.79\%, outperforming PROBE~\cite{probe} (70.60\%) and DINO-Det~\cite{dinodet} (76.50\%), despite using a simpler architecture.
In particular, DINO-Det relies on DINOv3-7B~\cite{simeoni2025dinov3} as its image encoder, which introduces substantially higher computational cost and may limit deployment practicality.
More detailed results are provided in Tab.~\ref{tab:fake_compare_detail}.

\begin{table}[htbp]
    \centering
    \vspace{-4mm}
    \caption{Accuracy (\%) comparison on FakeBench. We report Fake Accuracy (F.Acc), Real Accuracy (R.Acc), and Average Accuracy (Avg.) to comprehensively evaluate detector performance and generalization capability.
    N.B.\ 2 and GPT-I.\ 2 denote Nano Banana 2 and GPT-Image 2, respectively.
    The best and second-best results are highlighted in \textbf{bold} and \uline{underline}, respectively.
    All methods are evaluated using the official checkpoints released by their authors.}
    \resizebox{1.0\textwidth}{!}{
    \begin{tabular}{lccccccccccccccc}
    \toprule
        \multirow{2}{*}{\textbf{Methods}}& \multicolumn{2}{c}{\textbf{SD3.5}} & \multicolumn{2}{c}{\textbf{FLUX.1}} & \multicolumn{2}{c}{\textbf{FLUX.2}} & \multicolumn{2}{c}{\textbf{Z-Image}} 
        & \multicolumn{2}{c}{\textbf{Qwen-Image}} & \multicolumn{2}{c}{\textbf{N.B. 2}} & \multicolumn{2}{c}{\textbf{GPT-I. 2}}
        &\multirow{2}{*}{\textbf{Avg.}}
        \\
        \cmidrule(lr){2-3} \cmidrule(lr){4-5} \cmidrule(lr){6-7} \cmidrule(lr){8-9} \cmidrule(lr){10-11} \cmidrule(lr){12-13} \cmidrule(lr){14-15}
             &R.Acc  &F.Acc &R.Acc  &F.Acc  &R.Acc  &F.Acc &R.Acc  &F.Acc  &R.Acc  &F.Acc &R.Acc  &F.Acc  &R.Acc  &F.Acc & 
             \\
        \midrule
        UniFD &98.90   &2.70 &98.72    &0.19 &98.83    &0.50 &98.80  &1.03 &98.63   &0.51 &99.59   &0.00 &98.54   &0.00 &49.78 \\ 
        LOTA &8.35 &92.04 &8.25 &91.28 &8.34 &88.70 &8.43 &91.04 &8.71 &88.54 &8.57 &90.20 &8.76 &95.62  &49.77\\
        DRCT &85.16    &29.05 &85.04    &21.78 &84.72   &27.65 &85.26    &41.82 &85.17   &30.12 &82.45   &22.86 &86.50   &22.26 &56.42\\
        FatFormer  &99.64    &2.12 &99.55    &0.04 &99.54   &0.22 &99.59 &0.94 &99.54    &0.49 &100.00  &0.00 &99.64   &12.04 &50.95\\
        NPR &61.85   &62.66 &61.59    &79.09 &60.40  &66.24 &61.55 &73.95 &60.94    &80.88 &63.67   &59.18 &62.04   &91.24 &67.52\\
        SAFE &94.94   &2.48 &95.02    &3.15 & 95.23   &2.41 &94.68  &1.88 &94.91   &3.88 &97.14   &5.31 &94.16  &77.74 &54.49\\
        AIDE &94.00    &13.29 &94.29    &20.40 &93.88   &12.15 &94.16 &21.95 &94.22   &22.63 & 95.92  &5.71 &94.89   &15.33 &55.20\\
        Effort  &38.74 &70.04 &39.74 &74.31 &38.30  &69.16 &39.45 &82.49  &39.09  &88.76 &40.00 &85.31  &40.51  &97.08 &60.21\\
        ForgeLens  &99.96  &0.19 &99.93  &0.66 &99.01   &25.13 &99.92   &0.23 &99.91  &7.37 &100.00  &0.00 &100.00  &98.91 &59.37\\
        LTD &99.77  &2.64 &99.74  &1.36 &99.77  &2.48 &99.72       &5.06 &99.77  &4.88 &100.00  &0.41 &100.00  &68.98 &56.04\\
        Aligned & 99.81    &49.30 &99.81    &16.27 &99.84   &3.01 &99.86    &12.12 &99.85    &21.39 &100.00      &42.04 &99.27   &10.58 &60.94\\
         B-Free &89.22   &55.74 &88.96    &8.49 &90.01   &5.35 &89.06    &37.28 &89.39   &2.78 &91.02    &0.82 &89.05    &0.73 &52.70 \\
         PROBE  &87.67 &96.65 &87.85 &19.18 &88.46 &28.66 &87.58 &71.62 &88.13 &77.43 &89.80 &42.45 &86.86 &36.13 &70.60\\
         OmniAID &88.87 &49.04 &88.69 &37.20 &88.95 &16.40 &88.12 &39.57 &88.69 &45.43 &89.80 &54.87 &89.05 &29.19 &63.84\\
         DDA &86.19    &76.34 &85.81    &30.97 &86.92   &22.70 &85.99    &63.12 &86.44  &56.47 &86.12 &45.31 &84.31    &51.09 &67.70\\
         DINO-Det & 94.11 &73.86 &93.96  & 86.82 & 94.24  &51.48 &93.88    &57.26 & 94.23    &58.92 &94.69  &29.39 &93.80   &54.38 &\uline{76.50}\\
         \midrule
         \rowcolor{mygray}\textbf{FPD~(Ours)} &91.23    &81.28 &93.25   &88.93 &91.11 &68.81 &90.08   &76.35 &89.32 &77.33 &88.54 &32.99 &88.89 &59.03 &\textbf{79.79}\\
         \bottomrule
    \end{tabular}
    }
    \label{tab:fake_compare}
\end{table}

\begin{table}[htbp]
    \centering
     \vspace{-6mm}
    \caption{Average / Balanced accuracy~(\%) comparison on ManipulationBench. FLUX$_{\textrm{Random}}$ and FLUX$_{\textrm{Object}}$ denote FLUX-Fill using random and object masks, respectively.
    Qwen-Edit and Step-Edit are abbreviations for Qwen-Image-Edit and Step1X-Edit-v1p2, respectively.}
    \label{tab:mau_compare}
    \resizebox{1.0\textwidth}{!}{
    \begin{tabular}{lccccccccc}
    \toprule
        \textbf{Methods}& \textbf{FLUX$_\textrm{Random}$} & \textbf{FLUX$_\textrm{Object}$}  & \textbf{FLUX.2-klein} & \textbf{Qwen-Edit} 
        & \textbf{Step-Edit} & \textbf{N.B. 2} & \textbf{GPT-I. 2}
        &\textbf{Avg. / B.Avg.}
        \\
        \midrule
            
        UniFD  &50.16 / 50.16  &39.02 / 50.83   &38.43 / 50.72  &39.61 / 50.39  &39.10 / 50.72 &37.64 / 50.26 &37.45 / 50.23 &40.20 / 50.47  \\ 
        LOTA &49.94 / 49.94 &60.44 / 49.93 &60.67 / 49.72 &59.47 / 49.78 &60.51 / 50.14 &57.12 / 47.43 &63.70 / 52.34 &58.83 / 49.89 \\
        DRCT &51.00 / 51.00  & 42.51 / 50.86  &41.60 / 50.56  &44.79 / 52.17 & 41.31 / 50.13 &49.92 / 57.21 &57.82 / 63.99 &46.99 / 53.70 \\
        FatFormer  &49.86 / 49.86  &37.90 / 49.95  &37.56 / 50.11  &39.06 / 50.05 &38.36 / 50.26 &37.32 / 50.00 &45.91 / 56.96  &40.85 / 51.02 \\
        NPR  &45.09 / 45.09 &40.99 / 46.02 &44.98 / 49.81  &52.40 / 55.17 &44.00 / 48.77 &52.70 / 55.53 &81.92 / 80.33  &51.72 / 54.38\\
        SAFE  & 52.22 / 52.22 &41.63 / 51.94  &39.00 / 49.99  &38.99 / 48.96 &37.40 / 48.26 & 37.97 / 49.46 &84.94 / 86.67 &47.45 / 55.35\\
        AIDE  &51.13 / 51.13  &39.86 / 50.65  &38.16 / 49.63 &50.45 / 58.63 &48.31 / 57.54 &43.54 / 54.43 &51.65 / 60.43 &46.15 / 54.63\\
        Effort &45.60 / 45.60 &46.81 / 45.70 &52.62 / 50.67 &59.98 / 56.98 &56.05 / 53.52 &64.16 / 60.85 &75.90 / 68.99 &57.30 / 54.61\\
        ForgeLens &50.01 / 50.01 &37.85 / 49.98 &37.46 / 50.05 &41.25 / 51.89 &39.41 / 51.17 &41.41 / 53.17 &91.82 / 93.49 &48.45 / 57.10\\
        LTD &49.88 / 49.88 &37.76 / 49.88 &37.37 / 49.97 &39.00 / 50.04 &37.95 / 49.98 &37.48 / 50.04 &66.57 / 73.40 &43.71 / 53.31\\
        Aligned  &50.51 / 50.51 &38.46 / 50.44  &37.94 / 50.42   &67.27 / 73.18 &38.51 / 50.43  &55.16 / 64.14 & 45.91 / 56.96 &47.68 / 56.58\\
         B-Free   &62.36 / 62.36   &57.67 / 63.97  &57.12 / 63.57  &77.92 / 79.95 &58.59 / 64.61 &34.70 / 45.43 &42.47 / 51.86 &55.83 / 61.67\\
         PROBE &61.67 / 61.67 &56.58 / 62.93 &59.58 / 65.46 &92.10 / 91.51 &55.12 / 61.83 &65.47 / 69.61 &80.34 / 82.47 &67.26 / 70.78\\
         OmniAID &52.66 / 52.66 &44.34 / 53.36 &47.75 / 56.40 &88.10 / 88.50 &54.57 / 61.46 &67.32 / 71.82 &56.55 / 63.08 &58.75 / 63.89\\
         DDA  &54.54 / 54.54 & 46.52 / 54.82  &55.51 / 62.29  &89.44 / 89.20  &56.77 / 62.88 &56.63 / 62.48 &80.20 / 81.96 &62.80 / 66.88\\
         DINO-Det &52.22 / 52.22  &42.41 / 52.97  &43.35 / 53.99  & 62.95 / 69.01 &45.69 / 55.53 &40.43 / 52.04 &65.14 / 71.55 &50.31 / 58.18\\
         \midrule
         \rowcolor{mygray}\textbf{FPD~(Ours)} &53.68 / 53.68 &46.01 / 54.09 & 49.89 / 58.00 &79.02 / 80.47   &54.17 / 60.70  &46.25 / 53.22 &82.42 / 83.71 &58.77 / 63.41
         \\
         \bottomrule
    \end{tabular}
    }
   
\end{table}

\textbf{ManipulationBench.}
Tab.~\ref{tab:mau_compare} reports accuracy and balanced accuracy on ManipulationBench. Unlike FakeBench~(Tab.~\ref{tab:fake_compare}), which evaluates fully synthesized images, ManipulationBench tests localized edits surrounded by largely real image content. 
It therefore provides an important and often more demanding test of sensitivity to spatially restricted synthetic traces.
On each FLUX-Fill subset, 15 of 17 detectors obtain balanced accuracy between 45\% and 55\%.
This level of detection difficulty is a widespread phenomenon: DINO-Det decreases from 76.50\% on FakeBench to 58.18\% balanced accuracy on ManipulationBench, whereas NPR decreases from 67.52\% to 54.38\%.
%

FPD achieves 58.77\% average accuracy and 63.41\% average balanced accuracy. Compared with the GenImage-trained checkpoints of ForgeLens (48.45\%/57.10\%) and AIDE (46.15\%/54.63\%) evaluated here, 
FPD improves mean balanced accuracy by 6.31\% and 8.78\%, respectively. 
This shows stronger aggregate performance on localized edits in the present evaluation, without attributing the gain to a particular design choice.
B-Free~\cite{bias-free} and PROBE~\cite{probe} perform comparatively well on the FLUX-Fill subsets, while PROBE and DDA~\cite{dda} achieve strong aggregate results. Possible contributing factors include B-Free's inpainted-image augmentation, PROBE's detector-guided hard examples, and DDA's pixel- and frequency-domain alignment with pixel mixup. 
We report class-specific results in Tab.~\ref{tab:mau_compare_detail}.

\begin{wraptable}{r}{0.5\textwidth}
    \centering
    \vspace{-6mm}
    \caption{Comparison on DailyBench using Qwen-Image-Train as training data.}
    \label{tab:daily_compare_qwen}
    \resizebox{0.5\textwidth}{!}{
    \begin{tabular}{lcccc}
    \toprule
        \multirow{2}{*}{\textbf{Methods}} &\multicolumn{2}{c}{\textbf{FakeBench}} & \multicolumn{2}{c}{\textbf{ManipulationBench}}
        \\
        \cmidrule(lr){2-3} \cmidrule(lr){4-5}
             &Avg.  &B.Avg. &Avg.  &B.Avg.
             \\
        \midrule
        UniFD &78.95   &78.95 &46.67  &53.02  \\ 
        LOTA  &55.03   &55.03 &45.86  &53.20  \\
        DRCT  &69.22  &69.22 &53.69  &60.80     \\
        NPR  &89.04   &89.04 &53.69  &61.03    \\
        AIDE  &76.56   &76.56 &56.62  &63.19    \\
        Effort  &91.23  &91.23 &46.17  &55.23  \\
        ForgeLens  &76.46 &76.46  &55.92  &63.09  \\
        LTD  &92.77 &92.77  &52.80  &60.55  \\
        Aligned  &81.96  &81.96  &56.33  &60.52   \\
         B-Free  &91.79 &91.79  &54.79  &62.16     \\
         OmniAID  &73.58   &73.58  &46.83  &52.38  \\
         DDA  &84.01   &84.01  &55.64  &62.37  \\
         DINO-Det  &92.58   &92.58  &50.80  &58.77    \\
         \midrule
         \rowcolor{mygray}\textbf{FPD~(Ours)}  &95.99    &95.99 &56.89  &63.21    \\
         \bottomrule
    \end{tabular}
    }
\end{wraptable}

\textbf{Training with Qwen-Image-Train Data.}
To evaluate the benefit of modern generator training data, we train each detector on Qwen-Image-Train using the authors' official implementations and report their DailyBench performance in Tab.~\ref{tab:daily_compare_qwen}.
Compared with the SDv1.4-trained results in Tab.~\ref{tab:fake_compare}, most methods obtain substantial gains on FakeBench; for example, UniFD improves from 49.78\% to 78.95\%, and LTD~\cite{ltd} increases from 56.04\% to 92.77\%.
FPD achieves the best FakeBench performance, reaching 95.99\% accuracy and outperforming DINO-Det and LTD by 3.41\% and 3.22\%, respectively.
These results show that up-to-date generator data substantially improves robustness to recent fully synthesized images.
However, this improvement does not transfer reliably to localized manipulation detection.
On ManipulationBench, most current methods remain around 50\% accuracy, with the best result reaching only 56.62\%.
This gap suggests that training on fully synthesized images alone is insufficient for localized AI-manipulated scenes, which require manipulation-aware training data or detector designs.
More detailed results are provided in Tabs.~\ref{tab:fake_compare_qwen_detail}, \ref{tab:mau_compare_qwen}, and \ref{tab:mau_compare_detail_qwen}.

\begin{table}[]
    \centering
     \vspace{-4mm}
    \caption{Balanced accuracy~(\%) comparison on In-the-wild Benchmarks using Qwen-Image-Train data for training.}
    \resizebox{0.9\textwidth}{!}{
    \begin{tabular}{lccccccccc}
    \toprule
        \multirow{2}{*}{\textbf{Methods}} & \multirow{2}{*}{\textbf{Chameleon}} &\multicolumn{3}{c}{\textbf{SynthWildX}} & \multicolumn{3}{c}{\textbf{WildRF}} &\multirow{2}{*}{\textbf{RRDataset}} 
        &\multirow{2}{*}{\textbf{Avg.}}
        \\
        \cmidrule(lr){3-5} \cmidrule(lr){6-8}
         & &DALLE3 &Firefly &Midj. &FB &Reddit &Twitter\\
         \midrule
         UniFD &62.24  &62.31  &\uline{69.51}  &61.53  &66.25  &51.67  &57.39  &70.27  &62.64 \\
         LOTA &53.37 &48.23 &48.65 &48.53 &46.56 &46.20 &45.72 &51.96 &48.65 \\
         DRCT &39.87  &40.15  &45.65  &41.10  &52.19  &40.93  &45.86  &52.57  &44.79\\
         NPR &67.53  &56.39  &56.02  &55.23  &58.44  &51.53  &49.57  &48.90  &55.45\\
         AIDE &57.43  &41.81  &50.96  &41.80  &58.75  &48.67  &43.05  &54.18  &49.58\\
          Effort &69.02  &57.56  &65.32  &62.45  &56.56  &47.73  &48.88  &57.33  &58.10\\
        ForgeLens &59.99  &46.55  &51.20  &46.82  &61.56  &56.93  &41.57  &60.76 &53.17\\
        LTD &80.32  &61.98  &58.28  &62.86  &53.75  &47.27  &52.79  &60.24  &59.68\\
         Aligned &59.00  &\uline{72.21}  &56.71  &58.28  &\uline{78.44}  &\uline{71.07}  &\uline{74.81}  &64.69  &66.90\\
         B-Free &76.47  &52.30  &52.85  &51.67  &55.00  &53.53 &49.42  &44.65  &54.48 \\
         OmniAID &61.77 &56.94 &56.75 &57.17 &45.31 &41.00 &45.40 &59.15 &52.93\\
         DDA &64.68  &38.36  &48.30  &39.03  &65.00  &50.00  &42.57  &54.91  &50.35\\
         DINO-Det &\uline{90.09}  &69.55   &67.12  &\uline{69.05}  &66.56  &68.73  &64.37 &\textbf{77.40}  &\uline{71.60}\\
         \midrule
         \rowcolor{mygray}\textbf{FPD} &\textbf{91.39}  &\textbf{92.08}  &\textbf{81.24} &\textbf{87.99} &\textbf{85.98} &\textbf{84.36} &\textbf{79.77} &\uline{77.15} &\textbf{85.00}\\
         \bottomrule
    \end{tabular}
    }
    \label{tab:inwild_qwen}
\end{table}
\textbf{Generalization on In-the-Wild Benchmarks.}
Qwen-Image-Train differs from existing in-the-wild benchmarks in both generator source and image distribution, making this evaluation a cross-generator and cross-domain generalization test.
As shown in Tab.~\ref{tab:inwild_qwen}, FPD achieves the best average balanced accuracy of 85.00\%, surpassing B-Free~\cite{bias-free} by 30.52\% and the runner-up DINO-Det~\cite{dinodet} by 13.40\%.
FPD also obtains the best results on seven out of eight evaluation subsets, including all SynthWildX and WildRF subsets.
These results suggest that FPD benefits more effectively from modern generator training data and transfers better to diverse in-the-wild image distributions than existing detectors.

\textbf{Robustness to Pixel Perturbations.}
Real-world images are frequently altered by post-processing operations during transmission, storage, and online re-sharing, which can weaken the visual artifacts used by AI-generated image detectors.
Such pixel-level perturbations are particularly challenging because they may remove subtle generative traces without changing the semantic content of the image.
To evaluate whether detectors remain reliable under these practical degradations, we test on FakeBench with two common perturbation families: JPEG Compression with quality factors $\mathrm{QF}\in\{90,75,50\}$ and Gaussian Blur with kernel size 5 and standard deviations $\sigma\in\{1,3,5\}$.

\begin{wraptable}{r}{0.6\textwidth}
    \centering
     \vspace{-6mm}
    \caption{Accuracy comparison on pixel perturbations using Qwen-Image-Train data for training.}
    \resizebox{0.6\textwidth}{!}{
    \begin{tabular}{lcccccc}
    \toprule
        \multirow{2}{*}{\textbf{Methods}} &\multicolumn{3}{c}{\textbf{JPEG Compression}} & \multicolumn{3}{c}{\textbf{Gaussian Blur}} 
        \\
        \cmidrule(lr){2-4} \cmidrule(lr){5-7}
         &QF=50 &QF=75 &QF=90 &$\sigma=1$ &$\sigma=3$ &$\sigma=5$\\
         \midrule
         UniFD &70.93 &78.80 &78.23 &80.24 &80.82 &80.84\\
         LOTA &54.81 &54.94  &55.33 &55.03 &54.64 &54.56\\
         DRCT &58.07 &60.27 &60.67 &59.39 &59.40 &59.52\\
         NPR &67.75 &77.74 &88.64 &89.79 &88.10 &87.49\\
         AIDE &69.77 &77.65 &77.74 &84.66 &81.87 &80.23\\
          Effort &\uline{85.67} &91.05 &91.51 &93.08 &\uline{92.84} &\uline{92.73}\\
        ForgeLens &54.91 &70.31 &83.73 &67.81 &52.15 &51.27\\
        LTD &78.29  &\uline{93.38} &93.16 &92.34 &78.85 &77.72\\
         Aligned &72.65 &80.15 &80.34 &84.07 &82.04 &81.19\\
         B-Free &85.13 &92.35 &91.83 &84.11 &67.21 &70.19\\
         OmniAID &73.37 &73.41 &72.49 &71.24 &70.77 &70.32\\
         DDA &57.08 &77.92 &83.89 &85.31 &82.71 &81.08\\
         DINO-Det &82.51 &90.77 &\uline{93.22} &\uline{93.40} &91.72 &91.19\\
         \midrule
         \rowcolor{mygray}\textbf{FPD} &\textbf{88.73} &\textbf{95.24} &\textbf{95.93} &\textbf{96.70} &\textbf{95.07} &\textbf{94.79}\\
         \bottomrule
    \end{tabular}
    }
   
    \label{tab:piexl_qwen}
\end{wraptable}

\textbf{Robustness under JPEG Compression.}
JPEG compression evaluates detector robustness to lossy transmission, where smaller quality factors indicate stronger compression.
As summarized in Tab.~\ref{tab:piexl_qwen}, detector performance generally decreases as the quality factor is reduced, suggesting that compression suppresses discriminative traces used to separate real and AI-generated images.
Nevertheless, FPD consistently achieves the best performance across all tested compression levels, indicating that its detection cues are less sensitive to compression-induced distribution shifts than those of competing detectors.
Detailed results for each compression setting are reported in Tabs.~\ref{tab:fake_jpeg50_qwen}, \ref{tab:fake_jpeg75_qwen}, and \ref{tab:fake_jpeg90_qwen}.

\textbf{Robustness under Gaussian Blur.} 
Gaussian blur evaluates detector robustness to image smoothing, where larger $\sigma$ values remove more high-frequency details from the input image.
As shown in Tab.~\ref{tab:piexl_qwen}, most detectors degrade as blur strength increases, suggesting that they rely on fine-grained artifacts that can be attenuated by smoothing.
In comparison, FPD maintains stronger performance across all blur settings, demonstrating better robustness to smoothing-induced distribution shifts.
The detailed results are provided in Tabs.~\ref{tab:fake_blur1_qwen}, \ref{tab:fake_blur3_qwen}, and \ref{tab:fake_blur5_qwen}.


\section{Conclusion}
This paper addresses the growing mismatch between existing AI-generated image detection benchmarks and the rapidly evolving capabilities of modern generative models.
We introduce \textbf{DailyBench}, a unified high-fidelity benchmark for evaluating AIID methods in both full-image synthesis and localized-manipulation settings.
DailyBench consists of two complementary subsets: \textbf{FakeBench}, which covers high-fidelity synthetic images generated by recent open-source and commercial text-to-image models, and \textbf{ManipulationBench}, which targets object-level edits generated by image-conditional models.
Together, these subsets provide a more realistic and comprehensive testbed for measuring detector robustness under modern generation and editing pipelines.
We further propose \textbf{FPD}, a diagnostic-driven baseline that combines dual-pathway representation learning, a cascaded compression classifier, and a fake-preference sampler.
Extensive experiments show that state-of-the-art detectors suffer substantial performance degradation on DailyBench, revealing a clear gap between existing benchmark performance and real-world generalization.
These findings highlight the need for AIID methods that are robust across modern generation sources and sensitive to localized manipulations.

\textbf{Limitations.} DailyBench represents a snapshot of rapidly evolving generative technology rather than an exhaustive benchmark of all deployed models, and it does not cover every deployed generator and editing operation.
Its commercial-generator subsets are relatively small because of limited access, collection costs, and evolving proprietary systems. 
The resolution of the real images in this benchmark is constrained by the distribution of the source images from LAION, whereas the generated images span a variety of resolutions and aspect ratios.
The current focus is primarily on image inpainting, object removal, and object replacement; other editing operations and occasional failures in automated processing pipelines fall outside its scope.
Future versions will incorporate more generators, image sources, visual domains, and manipulation types.


%
\subsection*{AI use statement}

In this work, we used generative AI tools for polishing the writing and generating text prompts and fake images in DailyBench, as described in Section~\ref{sec_daily}.
We have not used generative AI tools for retrieval and discovery, research ideation or execution, drafting sections of the paper, or proving mathematical claims.
%
We have reviewed all AI-assisted work and verified that content polished by LLMs did not deviate from the original meaning, and we take responsibility for the final content of this work, including text, claims, or artifacts produced with the aid of generative AI.

\subsection*{Reproducibility Statement}

\textbf{Dataset Availability.}
To facilitate reproducibility, we provide detailed descriptions of DailyBench in Section~\ref{sec_daily}. 
Additional details supporting the reproduction of DailyBench is provided in Section~\ref {bench_supp}.
We will release DailyBench on the Hugging Face and ModelScope platforms.

\textbf{Method Training.} We detail our method in Section~\ref{sup_fpd} and provide training details in Section~\ref{sup_setting}.

\bibliography{iclr2027_conference}
\bibliographystyle{iclr2027_conference}

\newpage
\appendix

\section{Diagnostic-Driven Strong Baseline} \label{sup_fpd}

\subsection{Motivation and Overview}

\begin{figure}[htbp]
    \centering
    \includegraphics[width=1.0\linewidth]{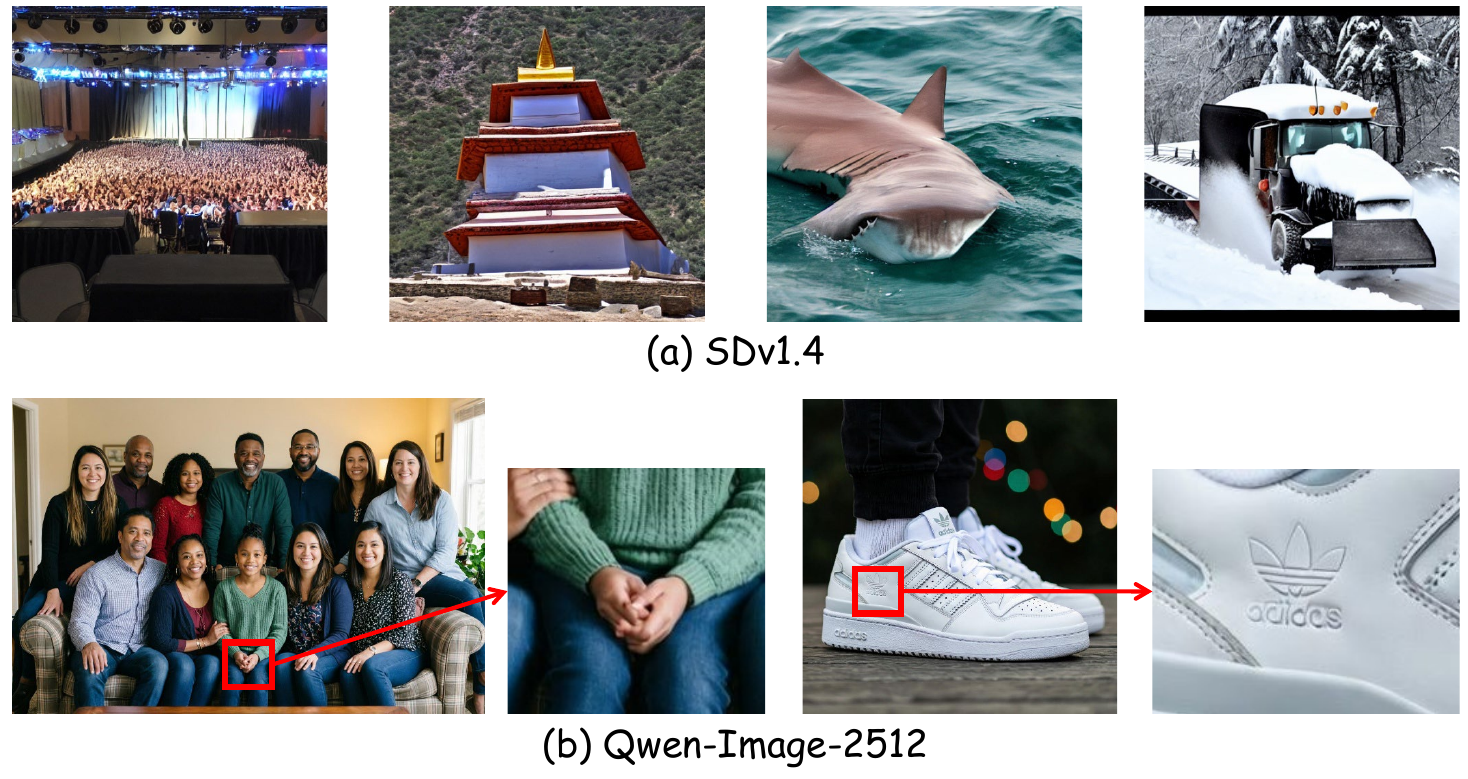}
    \caption{Illustrative comparison between fake images generated by SDv1.4 and Qwen-Image-2512. The SDv1.4 examples contain globally perceptible inconsistencies, whereas the Qwen-Image-2512 examples appear more globally coherent. This observation motivates combining a global semantic feature with a complementary summary of patch-token responses.}
    \label{fig:diagnostic_view}
\end{figure}

Our diagnostic analysis identifies two major obstacles to AI-generated image detection. First, the diagnostic evidence left by different generators varies in both form and granularity. 
Earlier generators, such as SDv1.4~\cite{sd}, often produce globally perceptible inconsistencies, whereas recent generators produce more coherent images whose synthetic traces may be subtler and expressed differently across image regions, as illustrated in Fig.~\ref{fig:diagnostic_view}. 
A detector relying on a single global representation may therefore fail to capture the diverse evidence produced by different generator families.

Addressing this first obstacle requires both an expressive feature representation and a classifier capable of integrating its complementary components. 
We first introduce the \emph{Dual-Pathway Feature} module, which combines the DINOv3 \texttt{[CLS]} token with a mean-pooled summary of the patch-token responses.
These pathways provide two distinct aggregations of the encoder features: one produced through the learned class token and the other computed directly from the patch tokens.
Because their concatenation is high-dimensional and may contain complex cross-pathway interactions, we further introduce the \emph{Cascaded Compression Classifier} (CCC), which progressively transforms and compresses the combined representation before binary classification.

The second obstacle is a systematic preference for the real class. As shown in Tab.~\ref{tab:fps_compare}, existing detectors generally achieve substantially lower accuracy on fake images than on real images.
Representation enhancement alone does not directly correct this asymmetric training behavior. We therefore introduce the \emph{Fake-Preference Sampler} (FPS), which constructs fake-dominant mini-batches and thereby increases the aggregate contribution of fake examples to the training objective.

\begin{figure}[htbp]
    \centering
    \includegraphics[width=1.0\linewidth]{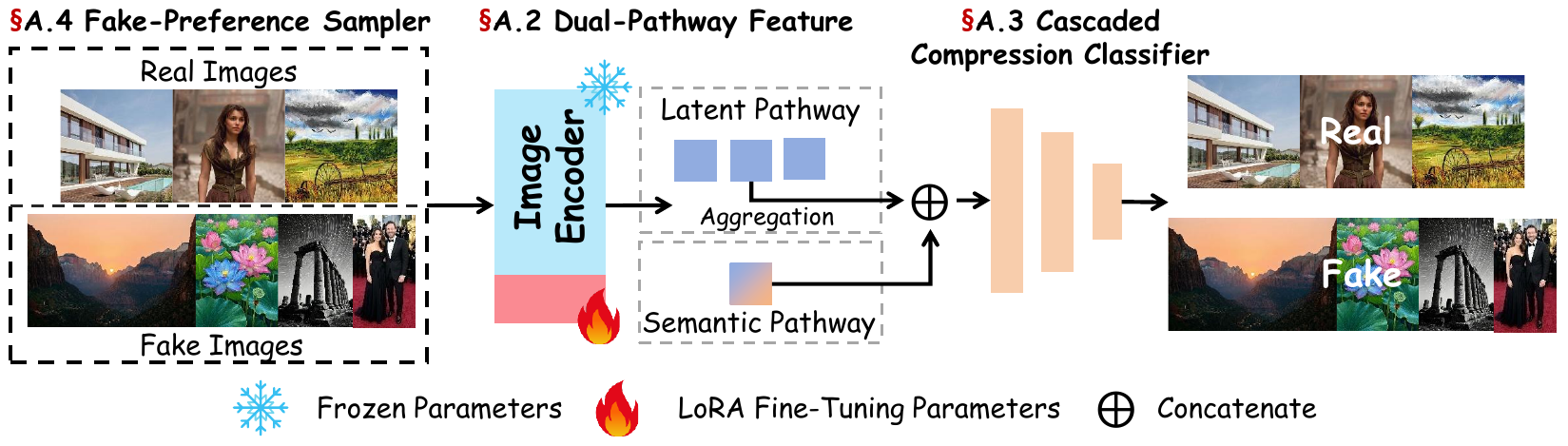}
    \caption{Overview of the proposed Fake-Preference Detector~(FPD).}
    \label{fig:method}
\end{figure}

Together, these three components form the \textbf{F}ake-\textbf{P}reference \textbf{D}etector (\textbf{FPD}), illustrated in Fig.~\ref{fig:method}. Given an input image, the Dual-Pathway Feature module constructs complementary global and patch-derived feature summaries. 
CCC then progressively integrates and compresses their concatenation to produce a binary prediction. During training, FPS adjusts the mini-batch composition to place greater optimization emphasis on the under-recognized fake class.

\subsection{Dual-Pathway Feature}
The first requirement identified by our diagnostic analysis is a representation that covers the heterogeneous evidence produced by different generators.
The DINOv3 \texttt{[CLS]} token provides a learned global summary of the image, but this single aggregation may not retain all information expressed across the patch-token responses. 
We therefore construct a dual-pathway feature representation that combines the \texttt{[CLS]} token with a direct summary of the patch tokens.

Given an RGB image $\mathbf{I}$, we use DINOv3~\cite{simeoni2025dinov3} to extract a sequence of visual tokens:
\begin{equation}
    \mathbf{E}
    =
    \mathcal{E}(\mathbf{I})
    =
    [\mathbf{e}_{\mathrm{cls}};\mathbf{e}_{1};\ldots;\mathbf{e}_{L}]
    \in
    \mathbb{R}^{(L+1)\times C},
\end{equation}
where $\mathbf{e}_{\mathrm{cls}}$ is the [CLS] token, $L$ is the number of image tokens, and $C$ is the feature dimension. 
We use the [CLS] token as the semantic-pathway feature $\mathbf{E}_{c}=\mathbf{e}_{\mathrm{cls}}\in\mathbb{R}^{C}$.
For the latent pathway, we aggregate the image tokens by average pooling $\mathbf{E}_{l}=\frac{1}{L}\sum_{i=1}^{L}\mathbf{e}_{i}\in \mathbb{R}^{C}$.
The final dual-pathway representation is obtained by concatenating the two features:
\begin{equation}
    \mathbf{E}_{f}
    =
    [\mathbf{E}_{c};\mathbf{E}_{l}]
    \in
    \mathbb{R}^{2C}.
\end{equation}
This representation allows the classifier to jointly exploit globally aggregated semantic information and complementary responses retained by the patch tokens. Consequently, it provides a broader feature basis for detecting both globally perceptible inconsistencies and subtler generation-related patterns.

\subsection{Cascaded Compression Classifier}
Constructing complementary features does not by itself determine how their information should be integrated for detection.
The concatenated dual-pathway representation is high-dimensional and may contain redundant components as well as nonlinear interactions between its two pathways.
We therefore introduce CCC to progressively transform and compress the representation before applying the final binary decision layer.

The Cascaded Compression Classifier progressively reduces the dimensionality of $\mathbf{E}_{f}$.
Let $\mathbf{h}_{0}=\mathbf{E}_{f}$ and let $d_{0}=2C>d_{1}>\cdots>d_{K}$ denote the dimensions of the successive stages. 
The $k$-th stage is formulated as
\begin{equation}
\mathbf{h}_{k}=\phi_{k}\left(\mathbf{W}_{k}\mathbf{h}_{k-1}+\mathbf{b}_{k}\right),
\qquad k=1,\ldots,K,
\end{equation}
where $\mathbf{W}_{k} \in \mathbb{R}^{d_{k} \times d_{k-1}}$ and $\mathbf{b}_{k}$ is learnable parameters and $\phi_{k}(\cdot)$ denotes the nonlinear transformation used at stage $k$.
In this paper, we set $K=3$ and $d_{k} = \frac{1}{2}d_{k-1}, k=\{1,2,3\}$.
The final prediction logit is computed as
\begin{equation}
z=\mathbf{w}_{o}^{\top}\mathbf{h}_{K}+b_{o},\qquad p(y=\mathrm{fake}\mid\mathbf{I})=\sigma(z).
\end{equation}

By introducing intermediate representations, CCC separates progressive feature compression from the final binary decision. 
Each stage reduces the dimensionality of the representation while preserving nonlinear interactions among the remaining features. 
This progressive design gives the classifier greater capacity to refine diagnostic-relevant evidence before producing the final prediction.

\subsection{Fake-Preference Sampler}
The preceding modules improve feature construction and classification, but they do not directly address the observed preference for predicting real images. 
To increase the influence of the under-recognized fake class during optimization, FPS constructs each training mini-batch with more fake images than real images. 
This sampling strategy changes the aggregate class contribution to the training loss while leaving the test distribution and evaluation protocol unchanged.

For each mini-batch of size $N$, FPS samples $N_{r}$ real images and $N_{f}$ fake images such that
\begin{equation}
    N=N_{r}+N_{f},
    \qquad
    N_{f}>N_{r}.
\end{equation}

This controlled training distribution increases the contribution of fake samples to the optimization objective and encourages the detector to model a broader range of fake-image characteristics.
Importantly, FPS is applied only during training and does not modify the class distribution or evaluation protocol of any test set.
\section{More Details on DailyBench}
\label{bench_supp}

\subsection{ManipulationBench}
We develop an automated pipeline to extract and localize the main objects in each image while filtering noisy data. The pipeline consists of two components:

\textbf{Object Extraction.}
We employ Qwen3-8B~\cite{yang2025qwen3} to extract object names from image captions using an in-context instruction: \textit{``Extract all object names in the text, one per line, without any description or explanation.''}

\textbf{Object Localization.}
We use Grounding-DINO~\cite{liu2024grounding} to generate bounding boxes for each object. To reduce noise and prevent excessively large manipulations, we discard objects whose bounding boxes occupy less than 2\% or more than 30\% of the image area.

\subsection{Statistics for Different Generators}
We report the per-generator sample statistics of FakeBench and ManipulationBench in Tab.~\ref{tab:detail_fake} and Tab.~\ref{tab:detail_mau}, respectively\footnote{We use the LAION-Aesthetics V2 image subset released by UniControl~\cite{qin2023unicontrol} to reduce real image storage and preprocessing cost.}.
In FakeBench, each AI-generated image is paired with a corresponding real image, enabling direct real/fake comparison under the same semantic prompt.
Therefore, the number of real and fake images is identical for each text-to-image generator.
In ManipulationBench, the sample count depends on the manipulation protocol.
Samples generated by FLUX-Fill$_\textrm{Random}$ follow a one-to-one correspondence with real images because each real image is edited once using a randomly selected region.
For the remaining manipulation generators, each real image can produce multiple manipulation scenarios, including object removal and object replacement.
This protocol results in more manipulated images than source real images for these generators.

\subsection{Benchmark Samples}
\begin{figure}
    \centering
    \includegraphics[width=0.75\linewidth]{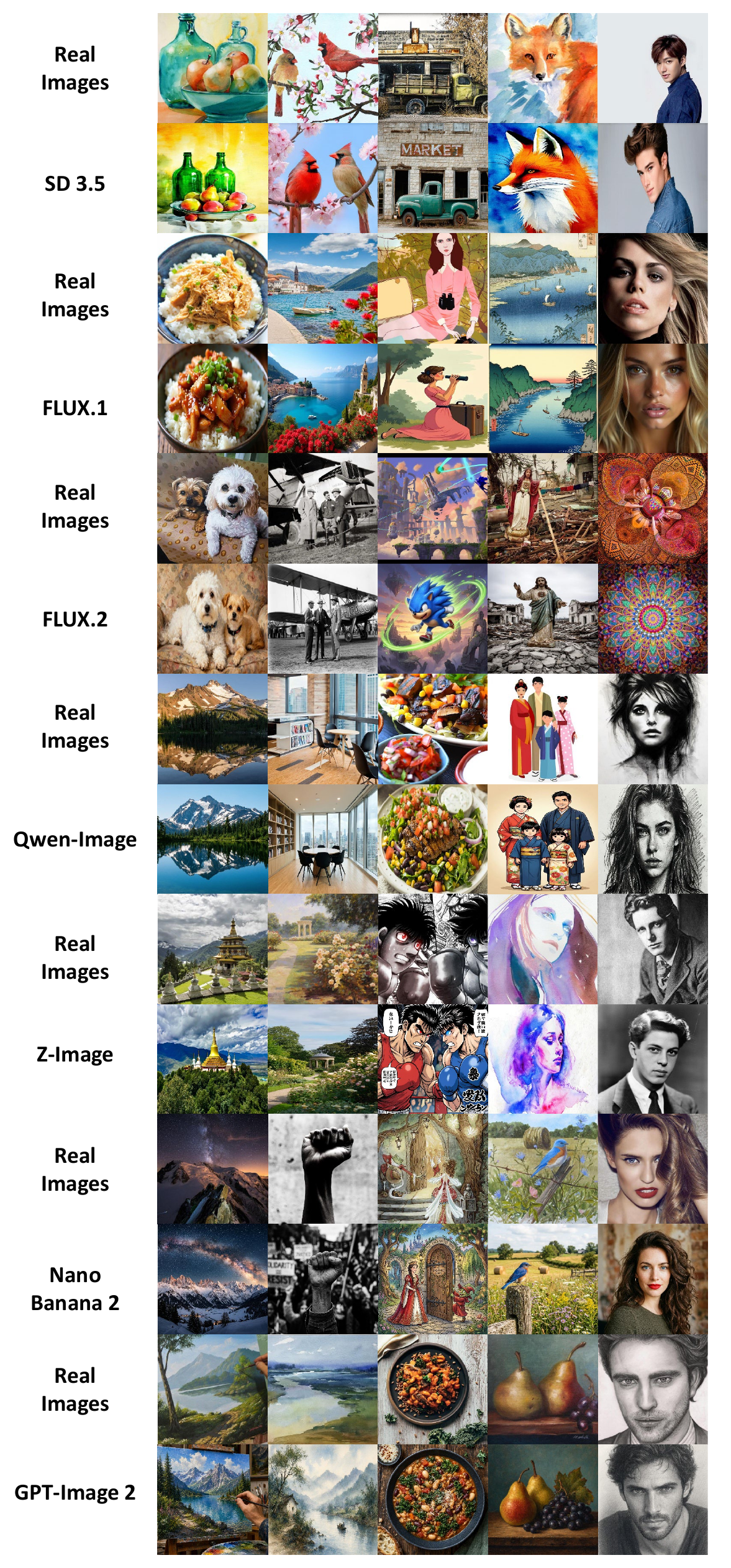}
    \caption{Examples from the FakeBench.}
    \label{fig:fake_view}
\end{figure}

\begin{figure}
    \centering
    \includegraphics[width=0.75\linewidth]{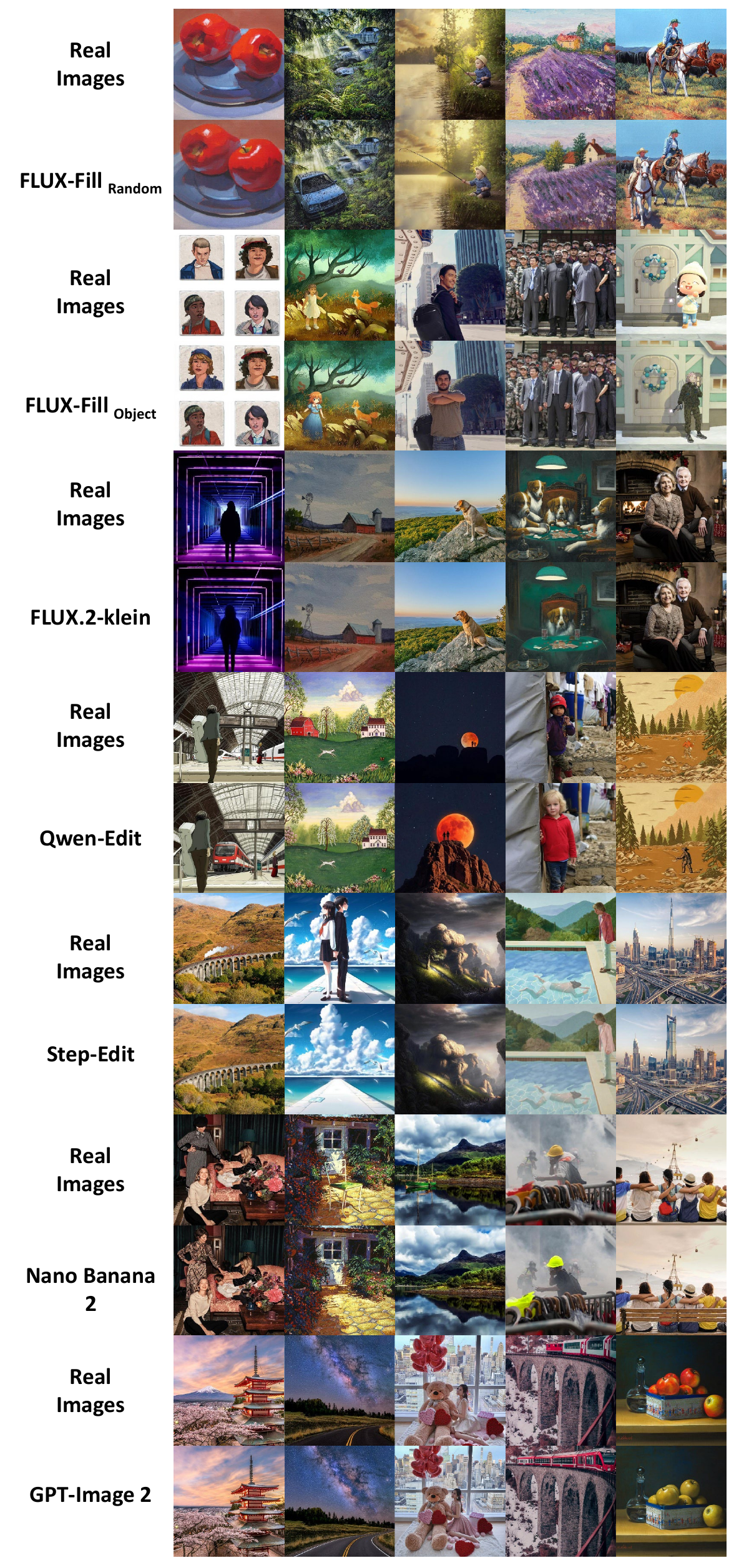}
    \caption{Examples from the ManipulationBench.}
    \label{fig:mau_view}
\end{figure}

We visualize random samples from FakeBench and ManipulationBench in Fig.~\ref{fig:fake_view} and Fig.~\ref{fig:mau_view}, respectively.
Fig.~\ref{fig:fake_view} shows that FakeBench contains diverse and visually realistic fake images across different generators.
Many generated images are difficult to distinguish from real images by human inspection alone, indicating the high visual quality of the benchmark.
Meanwhile, some generators still generate visible artifacts in specific cases; for example, the portrait in the fifth column contains noticeable artifacts for SD3.5 Large~\cite{sd3.5} and FLUX.1~\cite{flux2024}.
This observation aligns with the quantitative results in Tab.~\ref{tab:fake_compare}, where detectors obtain relatively higher accuracy on SD3.5 Large and FLUX.1 than on several more recent generators.

Fig.~\ref{fig:mau_view} shows representative localized manipulations from ManipulationBench.
These samples preserve most of the original real-image content while modifying only selected regions.
Compared with fully generated images in Fig.~\ref{fig:fake_view}, such manipulations contain fewer global synthetic artifacts and instead require detectors to recognize subtle object-level inconsistencies.
This qualitative evidence supports the design of ManipulationBench as a complementary evaluation setting for measuring robustness beyond full-image synthesis.

\begin{table}[]
    \centering
    \vspace{-4mm}
    \caption{Per-generator sample statistics of FakeBench. Each generated image is paired with a corresponding real image.}
    \resizebox{1.0\textwidth}{!}{
    \begin{tabular}{c|ccccccc}
    \toprule
        \textbf{Generator} &\textbf{SD3.5 Large} &\textbf{FLUX.1} &\textbf{FLUX.2} &\textbf{Qwen-Image} &\textbf{Z-Image} &\textbf{Nano Banana 2} &\textbf{GPT-Image 2} \\
        \midrule
        Real &19,398 &20,328 &12,456 &20,272 &19,321 &245 &274\\
        Fake &19,398 &20,328 &12,456 &20,272 &19,321 &245 &274\\
         \midrule
        Total &38,796 &40,656 &24,912 &40,544 &38,642 &490 &548\\
         \bottomrule
    \end{tabular}
    }
    \label{tab:detail_fake}
\end{table}

\begin{table}[]
    \centering
    \vspace{-4mm}
    \caption{Per-generator sample statistics of ManipulationBench. FLUX-Fill$_\textrm{Random}$ uses one edit per real image, while the remaining generators generate multiple object-level manipulation scenarios.}
    \resizebox{1.0\textwidth}{!}{
    \begin{tabular}{c|ccccccc}
    \toprule
        \textbf{Generator} & \textbf{FLUX-Fill$_\textrm{Random}$} & \textbf{FLUX-Fill$_\textrm{Object}$} & \textbf{FLUX.2-klein} &\textbf{Qwen-Edit} &\textbf{Step-Edit} &\textbf{Nano Banana 2} &\textbf{GPT-Image 2} \\
        \midrule
        Real &19,243 &4,714 &3,898 &4,403 &4,351 &228 &259\\
        Fake &19,243 &7,740 &6,533 &6,915 &7,121 &383 &438\\
         \midrule
        Total &38,486 &12,454 &10,431 &11,318 &11,472 &611 &697\\
         \bottomrule
    \end{tabular}
    }
    \label{tab:detail_mau}
\end{table}

\section{Supplementary Experimental and Analysis}
\subsection{Experiment Setting}  \label{sup_setting}
\textbf{Implementation Details.}
We use DINOv3-Large~\cite{simeoni2025dinov3} as the image encoder and fine-tune it with LoRA~\cite{hu2022lora} using rank $r=8$.
Unless otherwise specified, all methods are trained on the SDv1.4 subset of GenImage~\cite{genimage}.
We train FPD with AdamW, an initial learning rate of $5\times10^{-5}$, cosine annealing, and 10 training epochs.
All images are cropped to $448 \times 448$ during both training and inference.
Experiments are conducted on NVIDIA A100 GPUs.

\textbf{Evaluation Metrics.} Following prior work, we report Accuracy (Acc) and Average Precision (AP). AP evaluates ranking performance independently of a decision threshold, whereas Acc measures binary classification performance. For Acc, we set a standard 0.5 decision threshold without dataset-specific calibration or knowledge of class priors

\subsection{Ablation Studies}

\begin{table}[]
    \centering
    \vspace{-4mm}
    \caption{Component Ablation of the FPD on Chameleon.}
    \begin{tabular}{cccccc}
    \toprule
          CCC &DPF &FPS &R.Acc &F.Acc & Avg.  \\
         \midrule
         & & & 97.78 & 78.73 &89.61  \\
         \midrule
         \ding{51} & & & 95.40 &88.42 &92.41\\
          \ding{51} & \ding{51} & &96.34 &89.18 &\uline{93.27} \\
          \ding{51} & & \ding{51} &92.26 &92.22 &92.24\\
          \midrule
          \rowcolor{mygray}\ding{51} &\ding{51} & \ding{51} & 94.33 &93.46 &\textbf{93.96}\\
         \bottomrule
    \end{tabular}
    \label{tab:ablation}
\end{table}

\begin{figure}
    \centering
    \includegraphics[width=0.4\textwidth]{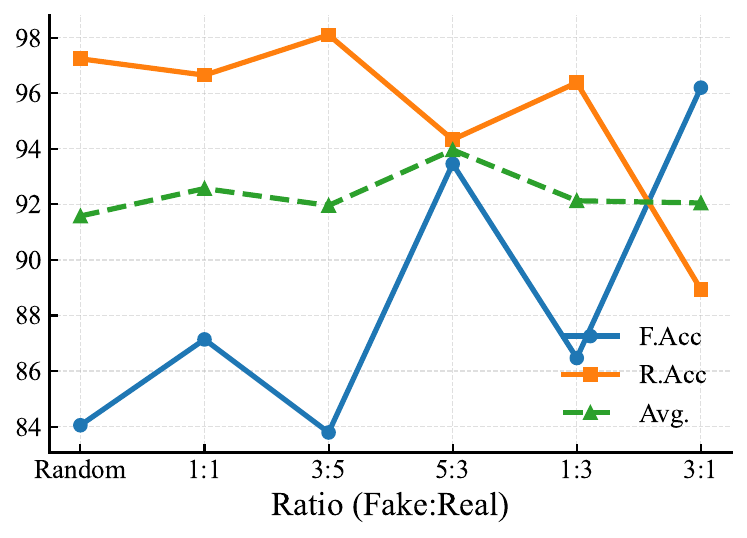}
    \caption{Ablation studies on fake-preference sampler.}
    \label{fig:fakesampler}
\end{figure}

\textbf{Component Ablation.}
Tab.~\ref{tab:ablation} evaluates the contribution of each FPD component.
We compare the DINOv3-Large baseline with four variants: (1) Cascaded Compression Classifier (CCC), (2) CCC with Dual-Pathway Feature (DPF), (3) CCC with Fake-Preference Sampler (FPS), and (4) the full FPD model combining CCC, DPF, and FPS.
The baseline shows a real-class bias, achieving 97.78\% accuracy on real images but only 78.73\% accuracy on fake images.
Adding CCC and DPF improves fake-image discrimination, showing that progressive feature compression and dual-pathway representations help capture artifact-relevant cues.
Further incorporating FPS produces more balanced performance, reaching 94.33\% accuracy on real images and 93.46\% accuracy on fake images.
These results verify that the three components are complementary and jointly improve detector robustness.

\textbf{Fake-Preference Sampler Ablation.}
We study the effect of different fake-to-real sampling ratios in Fig.~\ref{fig:fakesampler}.
Random sampling produces the largest gap between real-image and fake-image accuracy, indicating a strong bias toward predicting the real class.
This bias remains visible even when real and fake samples are equally sampled within each batch.
Increasing the proportion of fake samples reduces the gap and improves the balance between the two classes.
The best trade-off is achieved with a fake-to-real ratio of 5:3, which yields the most balanced performance on both real and fake image detection.

\begin{table}[]
    \centering
    \vspace{-4mm}
    \caption{Generalization experiments of FPS on Chameleon using Qwen-Image-Train as training data. R.Acc and F.Acc denote Real Accuracy and Fake Accuracy, respectively.}
     \resizebox{0.7\textwidth}{!}{
    \begin{tabular}{lcccc}
    \toprule
        Methods &Configurations &R.Acc  &F.Acc &Avg. 
             \\
        \midrule
        \multirow{2}{*}{AIDE~\cite{chameleon}} &w/o FPS  &87.65  &27.21  &61.72\\
         &w FPS  &87.43  &31.15 &63.29$_\mathrm{\uparrow\textcolor{green}{1.57}}$   \\
         \midrule
         \multirow{2}{*}{B-Free~\cite{bias-free}} &w/o FPS &97.77 &55.17  &79.50\\
         &w FPS  &98.41  &60.03   &81.94$_\mathrm{\uparrow\textcolor{green}{2.44}}$    \\
         \midrule
          \multirow{2}{*}{LTD~\cite{ltd}} &w/o FPS  &99.35  &61.29    &83.02   \\
         &w FPS  &99.37 &65.72 &84.93$_\mathrm{\uparrow\textcolor{green}{1.91}}$\\
         \bottomrule
    \end{tabular}
    }

    \label{tab:fps_compare}
\end{table}

\begin{table}[htbp]
    \centering
    \vspace{-4mm}
     \caption{Balanced accuracy~(\%) comparison on GenImage benchmark~\cite{genimage}. Our method is trained on the SDv1.4 training set, and the methods used for comparison employ their official checkpoints released by their authors.}
    \resizebox{1.0\textwidth}{!}{
    \begin{tabular}{lccccccccc}
    \toprule
        \textbf{Methods} &\textbf{ADM} &\textbf{VQDM} &\textbf{Midjourney} &\textbf{Wukong} &\textbf{GLIDE} & \textbf{BigGAN} &\textbf{SDv1.4} &\textbf{SDv1.5} &\textbf{Avg.}   \\ \midrule
         UniFD &58.10 &67.80 &91.50 &94.50 &73.40 &57.70 &96.40 &96.10 &79.50\\ 
        LOTA &99.40 &99.40 &91.30 &99.80 &98.20 &100.00 &99.90 &99.90 &98.50 \\
        DRCT &79.40 &90.00 &91.50 &94.70 &89.20 &81.70 &95.00 &94.40 &89.50\\
        FatFormer &76.46 &97.92 &93.43 &99.00 &88.71 &55.80 &100.00 &99.00 &88.93\\
        NPR &77.45 &84.94 &81.62 &97.67 &90.50 &84.20 &98.20 &98.68 &89.33\\
        SAFE &82.10 &96.30 &95.30 &98.20 &96.30 &97.80 &99.40 &99.30 &95.60\\
        AIDE &78.54 &80.26 &79.38 &98.65 &91.82 &66.89 &99.74 &99.76 &86.88\\
        Effort &78.70 &91.70 &82.40 &97.40 &93.30 &77.60 &99.80 &99.80 &91.10\\
        ForgeLens &80.65 &89.30 &64.18 &94.16 &94.92 &98.80 &95.87 &95.53 &89.18\\
        LTD &90.38 &92.53 &62.97 &96.33 &97.63 &98.61 &97.23 &97.24 &91.62 \\
        Aligned &52.40 &75.00 &97.50 &99.60 &57.60 &50.60 &99.70 &99.60 &79.00\\
         B-Free &79.80 &88.70 &95.70 &99.00 &85.30 &68.70 &98.80 &98.80 &89.30 \\
         PROBE  &91.00 &98.90 &98.80 &99.50 &92.10 &95.00 &99.40 &99.40 &96.80\\
         OmniAID &91.40 &97.30 &85.70 &98.10 &98.70 &98.70 &98.90 &98.80 &95.90\\
         DDA &94.30 &94.60 &94.40 &96.30 &93.10 &95.80 &97.30 &97.20 &95.50\\
         DINO-Det &87.82 &99.32 &93.85 &99.60 &97.39 &98.94 &99.79 &99.64 &96.77\\
         \midrule
         \rowcolor{mygray}\textbf{FPD~(Ours)} &77.30 &96.19 &93.66 &99.62  &92.47 &91.91 &99.69 &99.69 &93.81\\
         \bottomrule
    \end{tabular}
    }
   
    \label{tab:genimage}
\end{table}

\textbf{Generalization of Fake-Preference Sampler.}
To evaluate whether the fake-preference sampler generalizes across detector architectures, we integrate FPS into AIDE~\cite{chameleon}, B-Free~\cite{bias-free}, and LTD~\cite{ltd}, and report the results in Tab.~\ref{tab:fps_compare}.
FPS consistently improves fake-class accuracy for all three methods, with gains of 3.94\%, 4.86\%, and 4.43\% for AIDE, B-Free, and LTD, respectively.
These results suggest that real-class preference is a recurring issue in existing AIID models and demonstrate that FPS can serve as a general training strategy for improving fake-sample recognition across different detector architectures.

\subsection{Experiment Results with Outdated Generator Training Data}
\textbf{GenImage Results.} To demonstrate the generalization ability of FPD, we report the performance of FPD on the standard benchmark GenImage~\cite{genimage} in Tab.~\ref{tab:genimage}.
As shown in the table, several methods achieve average accuracies exceeding 90\%, suggesting that GenImage no longer poses a substantial challenge for current AI-generated image detectors.
In contrast, the performance of these state-of-the-art detectors degrades markedly on FakeBench~(Tab.~\ref{tab:fake_compare}) and ManipulationBench~(Tab.~\ref{tab:mau_compare}), both of which are constructed using modern generative models, with the most severe drop observed on ManipulationBench.
For example, LOTA~\cite{wang2025lota} achieves 98.50\% accuracy on GenImage, but this drops to 57.03\% on FakeBench and 57.08\% on ManipulationBench.
These results highlight the necessity of DailyBench as a more realistic and challenging benchmark for assessing the robustness and real-world generalization of AI-generated image detectors, particularly as modern generative models continue to narrow the perceptual gap between fake and real images.

\begin{table}[]
    \centering
    \vspace{-4mm}
    \caption{Balanced accuracy~(\%) comparison on In-the-Wild Benchmarks.}
    \resizebox{1.0\textwidth}{!}{
    \begin{tabular}{lccccccccc}
    \toprule
        \multirow{2}{*}{\textbf{Methods}} & \multirow{2}{*}{\textbf{Chameleon}} &\multicolumn{3}{c}{\textbf{SynthWildX}} & \multicolumn{3}{c}{\textbf{WildRF}} &\multirow{2}{*}{\textbf{RRDataset}} 
        &\multirow{2}{*}{\textbf{Avg.}}
        \\
        \cmidrule(lr){3-5} \cmidrule(lr){6-8}
         & &DALLE3 &Firefly &Midj. &FB &Reddit &Twitter\\
         \midrule
         UniFD &39.5 &52.5 &52.4 &50.7 &54.1 &55.1 &66.5 &51.1 &52.7\\
         LOTA &52.8 &50.0 &50.0 &49.8 &50.0 &79.9 &52.0 &64.8 &56.1\\
         DRCT &79.8 &85.9 &58.9 &90.5 &90.3 &66.8 &79.6 &59.2 &76.3\\
         FatFormer &57.8 &52.6 &56.6 &50.0 &52.5 &65.4 &39.7 &50.4 &53.1\\
         NPR &55.2 &62.9 &53.8 &63.0 &53.8 &53.7 &57.7 &48.3 &56.0\\
         SAFE &56.8 &48.8 &46.6 &48.7 &49.7 &49.2 &34.7 &49.3 &47.9\\
         AIDE &65.7 &66.4 &48.2 &66.4 &61.6 &66.3 &52.5 &57.6 &60.5\\
          Effort &67.3 &54.8 &54.9 &46.2 &51.8 &56.1 &49.7 &54.0 &54.3\\
        Forgelens &53.3 &49.9 &50.3 &50.1 &55.3 &75.8 &52.7 &63.2 &56.3\\
        LTD &52.6 &41.7 &67.9 &43.7 &64.0 &65.0 &48.0 &54.2 &54.6\\
         Aligned &61.3 &49.6 &53.9 &52.4 &48.4 &54.0 &40.6 &47.7 &50.9\\
         B-Free &78.3 &96.1 &92.3 &95.3 &95.6 &85.5 &96.7 &69.5 &88.6\\
         PROBE &86.6 &96.5 &96.4 &96.5 &98.4 &93.4 &99.0 &76.8 &92.9\\
         OmniAID &77.3 &75.6 &57.9 &57.8 &56.4 &64.2 &59.2 &64.6 &64.1\\
         DDA &83.5 &91.1 &84.7 &91.6 &85.3 &82.5 &89.3 &70.3 &84.7\\
         DINO-Det &93.6 &97.8 &90.8 &97.1 &93.8 &97.1 &95.8 &84.4 &\textbf{93.8}\\
         \midrule
         \rowcolor{mygray}\textbf{FPD} &93.9 &96.9 &88.1 &94.5 &98.1 &98.4 &98.7 &79.8 &\uline{93.5}\\
         \bottomrule
    \end{tabular}
    }

    \label{tab:inwild}
\end{table}

\textbf{Generalization on In-the-Wild Benchmarks.}
Real-world AI-generated image detection is challenging because images are often compressed, resized, reposted, or otherwise degraded during online dissemination.
As shown in Tab.~\ref{tab:inwild}, FPD achieves strong generalization across four in-the-wild benchmarks, obtaining an average balanced accuracy of 93.5\%.
FPD outperforms B-Free~\cite{bias-free} by 4.9\% and is only 0.3\% behind DINO-Det~\cite{dinodet}, the best-performing method in average balanced accuracy.
This is noteworthy because DINO-Det uses a 7B-parameter image encoder and blur augmentation during training, whereas FPD achieves comparable robustness with a simpler detector design.
Moreover, FPD obtains the best results on Chameleon, WildRF-Reddit, and WildRF-Twitter, demonstrating strong robustness to diverse online image distributions.

\textbf{More Detailed Results on FakeBench.}
Tabs.~\ref{tab:fake_compare_detail} and ~\ref{tab:fake_compare_ap} report the accuracy and average precision, respectively, on FakeBench.
Detector performance varies substantially across image generators, indicating that generalization remains sensitive to generator-specific characteristics. 
FPD achieves the highest average accuracy of 79.79\% and the average AP of 87.27\% among all evaluated methods. In terms of AP, it outperforms the recent PROBE~\cite{probe} and DINO-Det~\cite{dinodet} baselines by 14.61\% and 0.28\%, respectively. 
These results demonstrate that FPD transfers more consistently across fully synthesized images from different generator families.

\begin{table}[htbp]
    \centering
    \vspace{-4mm}
    \caption{Average accuracy (\%) comparison on FakeBench. All methods are evaluated using the official checkpoints released by their authors.}
    \resizebox{1.0\textwidth}{!}{
    \begin{tabular}{lccccccccccccccc}
    \toprule
        \textbf{Methods}& \textbf{SD3.5} &\textbf{FLUX.1} &\textbf{FLUX.2} & \textbf{Z-Image} 
        & \textbf{Qwen-Image} & \textbf{N.B. 2} & \textbf{GPT-I. 2}
        &\textbf{Avg.}
        \\
        \midrule
        UniFD &50.80 &49.45 &49.66 &49.91 &49.57 &49.80 &49.27 &49.78\\ 
        LOTA &50.19 &49.77 &48.52 &49.73 &48.62 &49.39 &52.19 &49.77\\
        DRCT &57.11 &53.41 &56.19 &63.54 &57.65 &52.65 &54.38 &56.42\\
        FatFormer  &50.88 &49.80 &49.88 &50.27 &50.01 &50.00 &55.84 &50.95\\
        NPR &62.25 &70.34 &63.32 &67.75 &70.91 &61.43 &76.64 &67.52\\
        SAFE &48.71 &49.09 &48.82 &48.28 &49.40 &51.22 &85.95 &54.49\\
        AIDE &53.64 &57.35 &53.02 &58.05 &58.42 &50.82 &55.11 &55.20\\
        Effort &54.39 &57.02 &53.73 &60.97 &63.93 &62.65 &68.80 &60.21\\
        Forgelens &50.07 &50.30 &62.07 &50.07 &53.64 &50.00 &99.45 &59.37\\
        LTD &51.20  &50.55 &51.12 &52.39 &52.32 &50.20 &84.49 &56.04\\
        Aligned &74.55 &58.04 &51.43 &55.99 &60.62 &71.02 &54.93 &60.94 \\
         B-Free &72.48 &48.72 &47.68 &63.17 &46.09 &45.92 &44.89 &52.70 \\
         PROBE &92.16 &53.51 &58.56 &79.60 &82.78 &66.12 &61.50 &70.60\\
         OmniAID &68.95 &62.94 &52.67 &63.84 &67.06 &72.35 &59.12 &63.84\\
         DDA &81.26 &58.39 &54.81 &74.56 &71.45 &65.71 &67.70 &67.70\\
         DINO-Det &83.99 &90.39 &72.86 &75.57 &76.58 &62.04 &74.09 &\uline{76.50}\\
         \midrule
         \rowcolor{mygray}\textbf{FPD~(Ours)} &86.26 &91.09  &79.96 &83.22 &83.32 &60.76 &73.96 &\textbf{79.79}\\
         \bottomrule
    \end{tabular}
    }
    \label{tab:fake_compare_detail}
    \vspace{-4mm}
\end{table}

\begin{table}[htbp]
    \centering
    \vspace{-4mm}
    \caption{Performance~(\%) comparison on FakeBench in AP. }
    \resizebox{1.0\textwidth}{!}{
    \begin{tabular}{lccccccccccccccc}
    \toprule
        \textbf{Methods}& \textbf{SD3.5} &\textbf{FLUX.1} &\textbf{FLUX.2} & \textbf{Z-Image} 
        & \textbf{Qwen-Image} & \textbf{N.B. 2} & \textbf{GPT-I. 2}
        &\textbf{Avg.}
        \\
        \midrule
        UniFD &61.54  &41.64 &47.48 &50.94 &46.30 &40.54 &39.89 &46.90\\ 
        LOTA &50.10 &49.73 &50.79 &50.05 &49.75 &49.80 &100.00 &57.17\\
        DRCT &61.58  &57.32 &60.89 &69.92  &62.82  &55.44 &57.59 &60.79\\
        FatFormer &55.24 &43.94  &44.46  &52.80 &50.97 &51.23 &81.29 &54.27 \\
        NPR &65.04 &76.89 &67.74 &72.13 &76.89 &62.70 &88.16 &72.79\\
        SAFE &47.69 &41.60 &46.17 &44.39 &49.90 &55.11 &95.75 &54.37 \\
       AIDE &63.23 &72.53 &61.21 &71.77 &74.44 &58.98 &64.45 &66.65\\
       Effort &53.11 &53.03 &54.07 &64.58 &67.41 &65.79 &88.33 &63.76 \\
        Forgelens &57.20 &54.67 &61.70 &55.02 &81.50 &65.80 &99.97 &67.98\\
        LTD &74.35 &70.20 &78.41 &81.00 &86.43 &63.98 &99.10 &79.07 \\
        Aligned &98.51 &92.52 &82.02 &88.62 &95.11 &97.51 &77.84 &90.30\\
         B-Free &83.86 &44.91 &43.78 &72.67 &39.64 &35.98 &35.51 &50.90\\
         PROBE  &97.95 &58.53 &63.35 &84.60 &69.31 &76.92 &58.01 &72.66\\
         OmniAID  &75.65 &67.90 &56.44 &70.16 &74.00 &77.47 &68.33 &69.99 \\
         DDA &86.53 &90.18 &59.07 &84.21 &81.71 &76.13 &78.56 &79.48\\
         DINO-Det &93.56 &96.47 &83.95 &87.62 &87.22 &71.86 &88.26 &86.99\\
         \midrule
        \rowcolor{mygray}\textbf{FPD~(Ours)} &94.26 &95.67 &87.84 &89.92 &90.04 &69.13 &84.05 &87.27\\
         \bottomrule
    \end{tabular}
    }

    \label{tab:fake_compare_ap}
\end{table}  

\begin{table}[htbp]
    \centering
    \vspace{-4mm}
    \caption{Detailed accuracy (\%) comparison on ManipulationBench.}
    \resizebox{1.0\textwidth}{!}{
    \begin{tabular}{lcccccccccccccc}
    \toprule
        \multirow{2}{*}{\textbf{Methods}}& \multicolumn{2}{c}{\textbf{FLUX$_\textrm{Random}$}} & \multicolumn{2}{c}{\textbf{FLUX$_\textrm{Object}$} } & \multicolumn{2}{c}{\textbf{FLUX.2-klein}} & \multicolumn{2}{c}{\textbf{Qwen-Edit}} 
        & \multicolumn{2}{c}{\textbf{Step-Edit}} & \multicolumn{2}{c}{\textbf{N.B. 2}} & \multicolumn{2}{c}{\textbf{GPT-I. 2}}
        \\
        \cmidrule(lr){2-3} \cmidrule(lr){4-5} \cmidrule(lr){6-7} \cmidrule(lr){8-9} \cmidrule(lr){10-11} \cmidrule(lr){12-13} \cmidrule(lr){14-15}
             &R.Acc  &F.Acc &R.Acc  &F.Acc  &R.Acc  &F.Acc &R.Acc  &F.Acc  &R.Acc  &F.Acc &R.Acc  &F.Acc &R.Acc  &F.Acc  
             \\
        \midrule
        UniFD & 98.92  &1.40 &99.43 &2.24 &99.36 &2.08 &98.98  &1.81 &98.85 &2.60 &100.00  &0.52  &100.00   &0.46  \\ 
        LOTA &8.33 &91.56 &6.66 &93.19 &6.39 &93.05 &6.11 &93.45 &7.19 &93.09 &9.21 &85.64 &8.11 &96.58\\
        DRCT &85.33  &16.68 &85.21  &16.50  &86.02  &15.09 &85.44   &18.90 &86.65 &13.61  &85.96  &28.46  &88.03  &39.95   \\
        FatFormer &99.58   &0.14 &99.55  &0.35  &99.77   &0.44 &99.55   &0.55 &99.54 &0.98 &100.00  &0.00  &100.00   &13.93 \\
        NPR &62.94 &27.24 &66.72 &25.32  &68.93  &30.69 &67.64  &42.70 &68.54 &29.01 &66.67  &44.39  &74.13    &86.53  \\
        SAFE &94.57 &9.86 &94.38 &9.50 &93.51  &6.47 &93.37   &4.56 &93.24 &3.27 &94.74  &4.18  &93.44    &79.91   \\
        AIDE &99.79  &1.23  &95.08  &6.23 &95.05   &4.21 & 95.48  &21.78  &95.79 &19.30  &97.37  &11.49  &94.59  &26.26  \\
        Effort &39.56  &51.63 &41.11 &50.28 &42.92 & 58.41 &43.42  &70.53 &43.02   &64.01 &47.81    &73.89 &42.08   &95.89\\
        Forgelens &99.94  &0.08 &99.89  &0.06 &99.90 &0.20 &99.84 &3.95 &99.89  &2.46 &99.56  &6.79 &100.00 &86.99\\
        LTD &99.73   &0.03  &99.72 &0.03 &99.85 &0.09 &99.80  &0.29 &99.77  &0.18 &99.56   &0.52 &100.00  &46.80\\
        Aligned &50.51 &50.51 &99.75   &1.14 &99.79   &1.04 &99.80 & 46.57 &99.79 &1.07 &99.56 &28.72  &100.00 &13.93  \\
         B-Free &89.76 &34.95 &89.90   &38.04 &89.12   &38.02 &89.12   &70.79 &89.54 &39.67  &87.72 &3.13  &88.42   &15.30    \\
         PROBE &87.95 &35.39 &89.03 &36.82 &88.71 &42.20 &88.85 &94.17 &89.66 &34.01 &85.96 &53.26 &90.73 &74.20\\
         OmniAID &88.74 &16.58 &90.45 &16.26 &90.61 &22.19 &90.30 &86.71 &89.97 &32.94 &89.47 &54.16 &89.65 &36.96\\
         DDA &85.72  &23.35 &89.01  &20.63 &89.12  &35.45 &88.12  &90.28 &88.19 &37.58  &85.53  &39.43  &88.80   &75.11  \\
         DINO-Det &94.23 &10.22 &96.41  &9.52 &96.13 &11.86 &96.28   &41.74  &96.25  &14.80 &97.81  &6.27  &96.53  &46.58  \\
         \midrule
         \rowcolor{mygray}\textbf{FPD~(Ours)} &91.60 &15.77 &87.42 &20.75  &90.21 &25.80 &87.01 &73.92  &87.84  &33.56  &81.82    &24.62  &88.97  &78.45   \\
         \bottomrule
    \end{tabular}
    }
    \label{tab:mau_compare_detail}
\end{table}

\textbf{More Detailed Results on ManipulationBench.}
ManipulationBench presents a substantially more difficult generalization setting than fully synthesized image detection. 
As shown in Table~\ref{tab:mau_compare_detail}, most detectors retain high real-image accuracy while their fake-image accuracy falls below 50\% on many manipulation sources. 
This pronounced imbalance indicates that locally manipulated images are frequently classified as real. A plausible explanation is that local editing preserves most of the natural image content, leaving only spatially restricted forensic traces. This explanation is consistent with the observed results but requires region-level analysis for direct validation.
B-Free~\cite{bias-free}, PROBE~\cite{probe}, and DDA~\cite{dda} achieve comparatively strong fake-image recognition on several manipulation sources. 
Their training strategies provide potentially relevant signals: B-Free incorporates inpainting-generated samples, PROBE generates detector-challenging examples through adversarial optimization, and DDA constructs augmented samples by combining real and fake image content. 
These results support the hypothesis that exposure to localized, composited, or hard-to-classify examples improves sensitivity to manipulated regions.
Table~\ref{tab:mau_compare_ap} provides a threshold-independent comparison using AP. PROBE achieves the highest mean AP of 85.31\%, followed by Aligned at 82.60\% and DDA at 80.63\%; FPD obtains 77.88\% mean AP.
It slightly exceeds B-Free and DINO-Det by 0.06\% and 2.63\%, respectively, but remains behind PROBE.

Overall, the detailed results reveal a clear gap between detecting fully synthesized images and detecting localized AI manipulations. 
While FPD achieves the strongest aggregate performance on FakeBench, ManipulationBench remains challenging for both FPD and existing baselines. Closing this gap will likely require manipulation-aware training signals that explicitly model the relationship between edited regions and preserved natural content.

\begin{table}[htbp]
    \centering
    \vspace{-4mm}
     \caption{Performance~(\%) comparison on ManipulationBench in AP.}
    \resizebox{1.0\textwidth}{!}{
    \begin{tabular}{lcccccccc}
    \toprule
        \textbf{Methods}& \textbf{FLUX$_\textrm{Random}$} & \textbf{FLUX$_\textrm{Object}$}  & \textbf{FLUX.2-klein} & \textbf{Qwen-Edit} 
        & \textbf{Step-Edit} & \textbf{N.B. 2} & \textbf{GPT-I. 2}
        &\textbf{Avg.}
        \\
        \midrule
        UniFD &53.96 &70.76 &71.00 &66.64 &69.19 &61.54 &60.65 &64.82 \\ 
        LOTA &49.74 &61.87 &62.27 &60.80 &61.77 &62.54 &100.00 &65.57\\
        DRCT &52.50 &63.61 &63.82 &64.28 &61.71 &71.52 &83.11 &65.79\\
        FatFormer &47.28 &64.07 &62.69 &59.75 &62.98 &63.21 &86.04 &63.71\\
        NPR &44.51 &57.38 &62.28 &65.79 &60.26 &67.91 &93.01 &64.45\\
        SAFE &57.02 &68.12 &63.03 &58.91 &57.82 &63.68 &96.39 &66.42\\
       AIDE &53.62 &64.21 &62.60 &80.04 &79.03 &76.16 &84.52 &71.45\\
       Effort  &44.46 &57.61 &63.71 &67.80 &65.04 &75.04  &89.83  &66.21 \\
        Forgelens &59.22 &69.17 &64.79 &85.79
        &82.56 &87.01 &99.53 &78.30 \\
        LTD &44.37 &58.61 &61.01 &66.48 &67.18 &70.94 &98.42 &66.72\\
        Aligned &72.28 &79.69 &71.60 &98.54 &73.13 &95.74 &87.28 &82.60\\
         B-Free &73.09 &82.84 &81.79 &92.62 &82.81 &52.84 &66.98 &76.13\\
         PROBE &71.09 &81.99 &83.30 &98.42 &79.41 &87.01 &95.96 &85.31\\
         OmniAID  &55.34 &68.19 &72.84 &96.26 &78.30 &85.18 &79.63 &76.53\\
         DDA &60.23 &72.54 &80.70 &97.34 &80.47 &80.11 &92.99 &80.63 \\
         DINO-Det &57.89 &71.37 &73.29 &88.56 &73.90 &69.41 &91.27 &75.09 \\
         \midrule
        \rowcolor{mygray}\textbf{FPD~(Ours)} &60.40 &72.01 &76.12 &92.11 &77.01 &72.59 &94.91 &77.88 \\
         \bottomrule
    \end{tabular}
    }
   
    \label{tab:mau_compare_ap}
\end{table}  

\begin{table}[htbp]
    \centering
    \vspace{-4mm}
    \caption{Accuracy~(\%) comparison on Infinity.}
    \resizebox{0.4\textwidth}{!}{
    \begin{tabular}{lccc}
    \toprule
        \multirow{2}{*}{\textbf{Methods}}& \multicolumn{3}{c}{\textbf{Infinity}}  
        \\
        \cmidrule(lr){2-4}
             &R.Acc  &F.Acc &Avg. 
             \\
        \midrule
        UniFD &98.70 &0.13 &49.42 \\ 
        LOTA &8.64 &96.35 &52.49\\
        DRCT &85.26 &57.67 &71.46  \\
        FatFormer &99.61 &2.59  &51.10  \\
        NPR &61.61&94.51  &78.06 \\
        SAFE & 94.66 &0.25 &47.46 \\
        AIDE &94.58 &4.06 &49.32  \\
        Effort &39.11 &94.11 &66.61\\
        Forgelens &99.90 &17.07 &58.48\\
        LTD &99.69 &15.41 &57.55\\
        Aligned &99.77 &69.23 &84.50  \\
         B-Free &89.72 &95.20 &92.46   \\
         PROBE &87.69 &98.68 &93.18\\
         OmniAID &88.55 &82.47 &85.51\\
         DDA &86.13 &97.96 &92.05  \\
         DINO-Det &93.81 &98.54 &\uline{96.17} \\
         \midrule
         \rowcolor{mygray}\textbf{FPD~(Ours)} &94.57 &98.69 &\textbf{96.63} \\
         \bottomrule
    \end{tabular}
    }
    
    \label{tab:infinity_compare}
\end{table}

\textbf{Experiment Results on Infinity.}
To further analyze detector behavior on autoregressive image generation models, we construct an additional test subset using images generated by the open-source autoregressive generator Infinity~\cite{han2025infinity}. The detailed results are reported in Tab.~\ref{tab:infinity_compare}.
Unlike the flow-based generators included in FakeBench, Infinity is substantially easier for existing detectors to identify. Most recent methods achieve very high fake-image accuracy, with many exceeding 95\% F.Acc.
FPD achieves 98.69\% fake-image accuracy and the best overall average accuracy of 96.63\%, while DINO-Det~\cite{dinodet} also reaches 96.17\% average accuracy.
These results suggest that Infinity-generated images contain more distinguishable generation traces than images generated by recent flow-based foundation models.
The performance gap among modern detectors is also smaller on Infinity than on the main FakeBench benchmark~(Tab.~\ref{tab:fake_compare_detail}).
Although some early methods still struggle due to poor fake-image classification, most recent detectors can reliably recognize Infinity-generated images.
This indicates that Infinity may introduce stronger or more consistent generation artifacts that can be detected without highly specialized forensic modeling.
One possible reason is that Infinity follows an autoregressive generation paradigm, which differs from the flow-based architectures used by recent high-fidelity generators such as FLUX.2~\cite{flux2} and Qwen-Image~\cite{qwenimage}.
As a result, its generated images may exhibit distinct statistical patterns that remain easier to separate from natural images.

From the perspective of benchmark construction, including Infinity in FakeBench would introduce a large number of relatively easy samples and could artificially inflate detector performance.
Therefore, we exclude it from the final benchmark composition.


%
\begin{table}[htbp]
    \centering
    \vspace{-4mm}
    \caption{Average accuracy (\%) on FakeBench using the constructed Qwen-Image-Train dataset for training. 
    }
    \resizebox{1.0\textwidth}{!}{
    \begin{tabular}{lccccccccccccccc}
    \toprule
        \textbf{Methods}& \textbf{SD3.5} &\textbf{FLUX.1} &\textbf{FLUX.2} & \textbf{Z-Image} 
        & \textbf{Qwen-Image} & \textbf{N.B. 2} & \textbf{GPT-I. 2}
        &\textbf{Avg.}
        \\
        \midrule
        UniFD &65.39 &81.50 &75.61 &69.42 &87.86 &86.94 &85.95 &78.95 \\ 
        LOTA &55.27 &55.18 &53.69 &53.55 &61.20 &54.90 &51.46 &55.03 \\
        DRCT &61.85 &57.29 &61.24 &63.42 &89.71 &64.90 &86.13 &69.22 \\
        NPR&81.60 &76.47 &86.46 &86.98 &98.80 &95.71 &97.26 &89.04 \\
       AIDE &64.13 &63.10 &65.33 &65.20 &97.85 &82.86 &97.45 &76.56 \\
       Effort &83.61 &88.06 &89.69 &84.66 &99.16 &96.94 &96.53 &91.23 \\
        Forgelens &69.38 &74.40 &62.07 &58.42 &99.41 &72.45 &99.09 &76.46 \\
        LTD &90.24   &95.55 &90.69 &90.57 &98.39 &94.53 &89.42 &\uline{92.77} \\
        Aligned &85.29 &79.09 &74.04 &77.90 &89.75  &77.35 &90.33 &81.96 \\
         B-Free &80.17 &83.39 &88.57 &90.46 &99.98 &100.00 &100.00 &91.79 \\
         OmniAID  &68.84 &76.92 &72.73 &69.79 &82.10 &73.46 &71.27 &73.58 \\
         DDA &77.91 &81.56 &67.58 &85.87 &97.16 &80.20 &97.81 &84.01 \\
         DINO-Det&88.31 &96.97 &88.60 &85.97 &98.41 &93.27 &96.53 &92.58 \\
         \midrule
        \rowcolor{mygray}\textbf{FPD~(Ours)} &90.62 &96.43  &93.00 &92.87 &99.99 &99.22 &99.83 &\textbf{95.99}\\
         \bottomrule
    \end{tabular}
    }
    \label{tab:fake_compare_qwen}
\end{table}  

\subsection{Experiment Results with Qwen-Image-Train Data}


\textbf{FakeBench.} Tabs.~\ref{tab:fake_compare_qwen}, \ref{tab:fake_compare_qwen_detail}, and \ref{tab:fake_compare_qwen_ap} report the overall accuracy, class-wise accuracy, and AP, respectively, on FakeBench.
FPD achieves the highest average accuracy of 95.99\%, outperforming the strongest competing methods, LTD and DINO-Det, by 3.22\% and 3.41\%, respectively. Across the seven test generators, FPD ranks first on SD3.5, FLUX.2, Z-Image, and Qwen-Image, and second on FLUX.1, Nano Banana 2, and GPT-Image 2. This consistent performance indicates that FPD generalizes more uniformly across generator families than the competing detectors.
Under an otherwise identical detector and evaluation protocol, replacing the GenImage SDv1.4 training set in Tab.~\ref{tab:fake_compare} with Qwen-Image-Train increases the average accuracy of FPD from 79.79\% to 95.99\%, an absolute improvement of 16.20\%.
The AP results exhibit a similar trend: FPD achieves a mean AP of 99.29\% and maintains an AP of at least 98\% on every evaluated generator, outperforming DINO-Det~\cite{dinodet} and B-Free~\cite{bias-free} by 1.09\% and 0.4\%, respectively. Thus, its advantage is preserved under a threshold-independent ranking metric.
These results suggest that training data from a more recent generator can provide cues that transfer more effectively to contemporary fully synthesized images.

\begin{table}[htbp]
    \centering
    \vspace{-4mm}
     \caption{Detailed accuracy (\%) comparison on FakeBench using Qwen-Image-Train data for training.}
    \resizebox{1.0\textwidth}{!}{
    \begin{tabular}{lcccccccccccccc}
    \toprule
        \multirow{2}{*}{\textbf{Methods}}& \multicolumn{2}{c}{\textbf{SD3.5}} & \multicolumn{2}{c}{\textbf{FLUX.1}} & \multicolumn{2}{c}{\textbf{FLUX.2}} & \multicolumn{2}{c}{\textbf{Z-Image}} 
        & \multicolumn{2}{c}{\textbf{Qwen-Image}} & \multicolumn{2}{c}{\textbf{N.B. 2}} & \multicolumn{2}{c}{\textbf{GPT-I. 2}}
        \\
        \cmidrule(lr){2-3} \cmidrule(lr){4-5} \cmidrule(lr){6-7} \cmidrule(lr){8-9} \cmidrule(lr){10-11} \cmidrule(lr){12-13} \cmidrule(lr){14-15}
             &R.Acc  &F.Acc &R.Acc  &F.Acc  &R.Acc  &F.Acc &R.Acc  &F.Acc  &R.Acc  &F.Acc &R.Acc  &F.Acc  &R.Acc  &F.Acc  
             \\
        \midrule
        UniFD &85.12  &45.65 &85.39  &77.60 &84.63  &66.59 &85.43  &53.42 &84.40  &91.32 &81.63   &92.24 &84.31 &87.59 \\ 
        LOTA &89.53 &21.01 &89.72 &20.65 &89.48 &17.89 &89.07 &18.02 &88.68 &33.73 &91.02 &18.78 &87.59 &15.33\\
        DRCT &96.25  &27.45 &98.86 &15.72 &95.33  &27.15 &95.12  &31.72 &94.97  &84.44 &94.69 &35.10 &95.62  &76.64\\
        NPR &98.08 &65.11 &97.92 &55.01 &97.69   &75.22 &97.74  &76.23 &97.71  &99.88  &96.73  &94.69 &95.99 &98.54 \\
        AIDE &94.72 &33.53 &96.01  &30.19 &96.36 &34.30 &96.10 &34.24 &96.01  &99.69 &95.51  &70.20 &96.72   &98.18 \\
        Effort  &99.53 &67.69 &99.51 &76.60 &99.68  &79.70 &99.52  &69.80  &99.52 &98.80  &99.59  &94.29   &99.64  &93.43 \\
        ForgeLens  &99.21 &39.56 &99.23  &49.58 &99.01 &25.13 &99.23 &17.61 &99.06  &99.76 &98.78  &46.12 &98.18 &100.00 \\
        LTD &97.70 &82.77 &95.75 &95.36 &97.80 &83.58 &97.70 &83.44 &98.70 &98.09 &100.00     &89.06 &98.91  &79.93 \\
        Aligned &80.34 &90.25 & 80.44 &77.74 &79.88 &68.20 &79.82  &75.98 &79.52 &99.99 &81.22 &73.47 &81.02  &99.64 \\
         B-Free &99.96 &60.38 &99.95  &66.84 &99.99   &77.15 &99.94 &80.98 &99.95  &100.00 &100.00    &100.00 &100.00    &100.00  \\
         OmniAID  &79.02 &58.67 &79.39 &74.44 &78.14 &67.33 &79.48 &60.10 &78.73 &85.47 &69.79 &77.14 &82.54 &60.00\\
         DDA &98.26  &57.57 &95.70   &67.41 &95.47 &39.68 &97.58  &74.16 &95.41 &98.90 &94.29  &66.12  &95.99  &99.64 \\
         DINO-Det &98.47  &78.15 &98.68  &95.26 &98.39  &78.82 &98.55 &73.38 &98.22  &98.59 &97.55  &88.98 &97.81  &95.26 \\
         \midrule
         \rowcolor{mygray}\textbf{FPD~(Ours)} &97.71 &83.53 &97.43 &95.43 &94.74 &91.26  &92.99  &92.74 &99.97  &100.00 &99.22  &99.22 &99.65 &100.00 \\
         \bottomrule
    \end{tabular}
    }
   
    \label{tab:fake_compare_qwen_detail}
\end{table}

\begin{table}[htbp]
    \centering
    \vspace{-4mm}
    \caption{Average precision (\%) on FakeBench using the constructed Qwen-Image-Train dataset for training. 
    }
    \resizebox{1.0\textwidth}{!}{
    \begin{tabular}{lccccccccccccccc}
    \toprule
        \textbf{Methods}& \textbf{SD3.5} &\textbf{FLUX.1} &\textbf{FLUX.2} & \textbf{Z-Image} 
        & \textbf{Qwen-Image} & \textbf{N.B. 2} & \textbf{GPT-I. 2}
        &\textbf{Avg.}
        \\
        \midrule
        UniFD &71.52 &86.78 &80.45 &75.56 &92.67 &88.97 &91.15 &83.87 \\
        LOTA &60.47 &59.95 &56.10 &57.40 &71.45 &59.48 &49.00 &59.12\\
        DRCT &61.55 &71.84 &67.68 &67.64 &80.91 &61.34 &60.73 &67.38\\
        NPR &94.63 &90.66 &96.01 &95.59 &99.95 &99.31  &99.88  &96.58\\
       AIDE &82.54 &87.01 &90.11 &83.92 &99.92 &94.39 &99.52 &91.05\\
       Effort &97.35 &99.04 &98.72 &97.22 &99.97 &99.95 &99.87 &98.87\\
        Forgelens &94.01 &95.42 &87.12 &86.30 &99.99 &95.60 &99.88 &94.05\\
        LTD &95.37 &97.60 &96.80 &96.16 &99.95 &99.82 &99.94 &97.95\\
        Aligned &93.85 &88.06 &83.90 &81.79 &99.98 &85.82 &99.94 &90.47\\
         B-Free &97.37 &98.42 &98.21 &98.23 &100.00 &100.00 &100.00 &98.89\\ 
         OmniAID  &76.53 &84.54 &79.76 &76.86 &89.75 &77.04 &78.83 &80.47 \\
         DDA &90.09  &92.80  &82.05 &83.02 &99.68 &92.01 &99.93 &91.36\\
         DINO-Det &96.74 &99.56 &97.09 &96.09 &99.88 &98.54 &99.54 &98.20\\
         \midrule
        \rowcolor{mygray}\textbf{FPD~(Ours)} &98.03 &99.70 &98.68 &98.72 &100.00 &99.98 &99.94 &99.29\\
         \bottomrule
    \end{tabular}
    }
    \label{tab:fake_compare_qwen_ap}
\end{table}  

\begin{table}[htbp]
    \centering
    \vspace{-4mm}
     \caption{Average / Balanced accuracy~(\%) comparison on ManipulationBench using Qwen-Image-Train data for training.}
    \resizebox{1.0\textwidth}{!}{
    \begin{tabular}{lccccccccc}
    \toprule
        \textbf{Methods}& \textbf{FLUX$_\textrm{Random}$} & \textbf{FLUX$_\textrm{Object}$}  & \textbf{FLUX.2-klein} & \textbf{Qwen-Edit} 
        & \textbf{Step-Edit} & \textbf{N.B. 2} & \textbf{GPT-I. 2}
        &\textbf{Avg. / B.Avg.}
        \\
        \midrule
        UniFD  &49.06 / 49.06  &37.78 / 46.85   &41.17 / 49.71    &52.00 / 57.96     &42.29 / 50.45     &48.45 / 55.59   &55.95 / 61.56  &46.67 / 53.02  \\ 
        LOTA &50.22 / 50.22 &40.56 / 50.22 &40.97 / 50.68 &45.33 / 53.32 &42.35 / 51.34 &45.66 / 54.88 &54.95 / 61.79 &45.86 / 53.20\\
        DRCT &51.74 / 51.74  &39.62 / 50.69   &41.10 / 52.20    &55.78 / 63.12     &46.56 / 56.06     &60.56 / 68.27    &80.49 / 83.53    &53.69 / 60.80 \\
        NPR  &49.59 / 49.59  &37.80 / 49.75   &37.47 / 49.81    &38.83 / 49.63    &37.74 / 49.54    & 77.58 / 81.67      &96.84 / 97.25   &53.69 / 61.03\\
        AIDE  &49.27 / 49.27  &37.44 / 48.92   &39.46 / 50.92    &72.38 / 76.70     &41.91 / 52.60     &58.92 / 66.79   &96.99 / 97.13   &56.62 / 63.19\\
        Effort &49.78 / 49.78 &37.70 / 49.80   &37.37 / 49.91   &38.90 / 49.93     &37.87 / 49.86    &49.43 / 59.66    &72.17 / 77.70    &46.17 / 55.23\\
        Forgelens &50.22 / 50.22  &37.86 / 49.80   &38.14 / 50.57 &75.15 / 79.63 &41.09 / 52.52    & 50.57 / 60.49    & 98.42 / 98.43     &55.92 / 63.09\\
        LTD &49.89 / 49.89 &37.75 / 49.84   &38.16 / 50.51    &51.10 / 59.91     &38.70 / 50.57     &71.85 / 77.37   &82.21 / 85.77  &52.80 / 60.55\\
        Aligned  &47.48 / 47.48  &39.14 / 47.66   &50.66 / 57.05    &58.45 / 62.82     &47.83 / 54.66     &57.94 / 63.16   & 92.83 / 90.82     &56.33 / 60.52\\
         B-Free   &49.98 / 49.98  &37.84 / 49.99   &37.37 / 49.99    &38.92 / 50.01     &37.93 / 50.00    &83.80 / 86.99   &97.70 / 98.17   &54.79 / 62.16\\
         OmniAID &49.57 / 49.57 &40.14 / 47.52 &42.93 / 50.43 &49.67 / 55.27 &42.90 / 50.06 &52.35 / 58.79 &50.28 / 55.07 &46.83 / 52.38\\
         DDA  &48.81 / 48.81  &37.53 / 49.11   &39.52 / 51.05    &64.45 / 70.24     &40.93 / 51.81     &63.18 / 70.27   &95.12 / 95.33    &55.64 / 62.37\\
         DINO-Det &50.92 / 50.92  &38.64 / 50.36   &38.99 / 50.96    &56.21 / 63.81     &42.92 / 53.68    &48.45 / 58.34    &79.48 / 83.36   &50.80 / 58.77\\
         \midrule
         \rowcolor{mygray}\textbf{FPD~(Ours)} &51.77 / 51.77  &38.69 / 49.43 &43.60 / 53.46    &63.56 / 69.31     &44.78 / 54.48    &60.94 / 68.27    &94.89 / 95.77    &56.89 / 63.21 
         \\
         \bottomrule
    \end{tabular}
    }
   
    \label{tab:mau_compare_qwen}
\end{table}

\textbf{ManipulationBench.}
Tabs.~\ref{tab:mau_compare_qwen}, \ref{tab:mau_compare_detail_qwen},  and \ref{tab:mau_compare_qwen_ap} report accuracy and AP on ManipulationBench when all detectors are trained on Qwen-Image-Train.
FPD achieves an overall accuracy of 56.89\% and a balanced accuracy of 63.21\%. 
These figures represent overall accuracy improvements of 1.25\% and 2.10\%, respectively, over the previously high-performing DDA~\cite{dda} and B-Free~\cite{bias-free} methods, which utilized training data resembling AI manipulations; this not only validates the effectiveness of the FPD method's design but also further underscores the importance of training data for the detection of AI-manipulated images.
%
%
%
%
The AP results provide a complementary, threshold-independent evaluation. FPD achieves a mean AP of 74.96\% and exceeds DDA and DINO-Det by 0.32\% and 1.97\%, respectively. 
Overall, these results indicate that localized AI manipulation remains challenging and motivate training signals that explicitly target edited-versus-preserved region inconsistencies.

Together with the FakeBench results, these findings expose a clear limitation of training exclusively on fully synthesized images. Although Qwen-Image-Train improves generalization to images generated entirely by recent generators, it remains insufficient to detect localized manipulations that preserve most of the original image content. 

\begin{table}[htbp]
    \centering
    \vspace{-4mm}
    \caption{Detailed performance~(\%) comparison on ManipulationBench using Qwen-Image-Train data for training.}
    \resizebox{1.0\textwidth}{!}{
    \begin{tabular}{lcccccccccccccc}
    \toprule
        \multirow{2}{*}{\textbf{Methods}}& \multicolumn{2}{c}{\textbf{FLUX$_\textrm{Random}$}} & \multicolumn{2}{c}{\textbf{FLUX$_\textrm{Object}$} } & \multicolumn{2}{c}{\textbf{FLUX.2-klein}} & \multicolumn{2}{c}{\textbf{Qwen-Edit}} 
        & \multicolumn{2}{c}{\textbf{Step-Edit}} & \multicolumn{2}{c}{\textbf{N.B. 2}} & \multicolumn{2}{c}{\textbf{GPT-I. 2}}
        \\
        \cmidrule(lr){2-3} \cmidrule(lr){4-5} \cmidrule(lr){6-7} \cmidrule(lr){8-9} \cmidrule(lr){10-11} \cmidrule(lr){12-13} \cmidrule(lr){14-15}
             &R.Acc  &F.Acc &R.Acc  &F.Acc  &R.Acc  &F.Acc &R.Acc  &F.Acc  &R.Acc  &F.Acc &R.Acc  &F.Acc &R.Acc  &F.Acc  
             \\
        \midrule
        UniFD &84.67 &13.45  &84.17 &9.52   &83.53 &15.89  &84.85 &31.08   &84.23  &16.67  &83.77  &27.42    &83.40   &39.73      \\ 
        LOTA &89.53 &10.92 &90.01 &10.44 &89.10 &12.26 &89.33 &17.31 &88.58 &14.10 &91.23 &18.54 &88.42 &35.16\\
        DRCT &96.73 &6.76  &96.27  &5.12   &96.13  &8.27   &96.18 &30.05   &95.40 &16.71  &98.68 &37.86    &95.37  &71.69      \\
        NPR &98.11 &1.08  &98.92 &0.58  &98.67 &0.95  &98.27 &0.98   &98.37 &0.70  &97.81  &65.54    &98.84   &95.66    \\
        AIDE &96.37 &2.17  &96.16 &1.68  &96.31  &5.54  &96.14 &57.25   &96.87  &8.33  &97.81 &35.77    &97.68  &96.58     \\
        Effort &99.55  &0.02  &99.60 &0.00  &99.54 &0.28  &99.64 &0.23   &99.52 &0.21  &100.00 &19.32    &99.23 &56.16    \\
        Forgelens &99.32  &1.12  &98.96 &0.65   &99.79  &1.35  &99.80 &59.46   &99.86  &5.18  &99.56   &21.41   &98.46 &98.40   \\
        LTD &99.69   &0.08  &99.62 &0.06  &99.44 &1.59  &99.64    &20.19   &99.72 &1.42 & 77.37 &99.12      &99.61  &71.92   \\
        Aligned &80.61 &14.34  &82.73 &12.60  &82.35 &31.75  &82.56 &43.09    &82.97  &26.36  &83.77  &42.55    &83.01  &98.63      \\
         B-Free &99.97 &0.00  &99.96 &0.01  &99.97 &0.02  &99.95  &0.06   &99.98 &0.01  &99.56 &74.41    &100.00  &96.35       \\
         OmniAID  &78.92 &20.21 &77.87 &17.16 &80.15 &20.71 &80.51 &30.03 &79.71 &20.41 &84.34 &34.24 &78.07 &33.86\\
         DDA &95.89  &1.73  &96.78  &1.45  &96.72  &5.39   &96.30 &44.18   &96.87  &6.74  &98.25 &42.30    &96.14 &94.52     \\
         DINO-Det &98.62 &3.22  &98.62 &2.11   &98.36 &3.57   &98.07  &29.56   &98.23  &9.13  &97.37  &19.32    &98.46 &68.26     \\
         \midrule
         \rowcolor{mygray}\textbf{FPD~(Ours)} &97.63 &5.92 &93.70   &5.16  &92.51 &14.42  &95.24 &43.38    &94.79 &14.18  &98.35 &38.19    &99.24 &92.31     \\
         \bottomrule
    \end{tabular}
    }
    \label{tab:mau_compare_detail_qwen}
\end{table}

\begin{table}[htbp]
    \centering
    \vspace{-4mm}
    \caption{Average precision~(\%) comparison on ManipulationBench using Qwen-Image-Train data for training.}
    \resizebox{1.0\textwidth}{!}{
    \begin{tabular}{lccccccccc}
    \toprule
        \textbf{Methods}& \textbf{FLUX$_\textrm{Random}$} & \textbf{FLUX$_\textrm{Object}$}  & \textbf{FLUX.2-klein} & \textbf{Qwen-Edit} 
        & \textbf{Step-Edit} & \textbf{N.B. 2} & \textbf{GPT-I. 2}
        &\textbf{Avg.}
        \\
        \midrule
        UniFD &49.59 &57.73 &62.53 &71.82 &62.39 &70.16 &75.38 &64.23  \\ 
        LOTA  &53.11 &64.49 &64.50 &69.68 &67.62 &73.20 &76.44 &67.01\\
        DRCT &51.94 &63.05 &68.40 &78.49 &72.28 &69.90 &71.23 &67.89\\
        NPR  &38.48 &49.90 &56.77 &53.01 &52.62 &95.01 &99.52 &63.62\\
        AIDE &45.94 &55.51 &66.85 &94.68 &72.11 &91.34 &99.69 &75.16\\
        Effort &51.39 &58.67 &61.92 &68.78 &61.48 &92.22 &97.71 &70.31\\
        Forgelens &69.19 &72.72 &76.40 &98.91 &90.93  &97.57 &99.38 &86.44 \\
        LTD  &42.30 &51.53 &62.04 &84.19 &64.92 &97.23 &99.59 &71.69\\
        Aligned &48.62 &60.13 &72.16 &99.68 &76.08 &90.96 &99.74 &78.19  \\
         B-Free &54.79 &68.61 &67.40 &69.30 &71.34 &96.04 &99.57 &84.67 \\
         OmniAID &49.48 &59.01 &63.05 &67.75 &62.12 &75.07 &70.47 &63.85\\
         DDA  &45.26 &58.77 &67.80 &91.23 &69.02 &91.01 &99.41 &74.64\\
         DINO-Det &53.62 &61.17 &64.18 &81.75 &69.67 &84.15 &96.42 &72.99\\
         \midrule
         \rowcolor{mygray}\textbf{FPD~(Ours)} &51.84 &59.07 &67.37 &84.07 &70.15 &92.88 &99.33 &74.96
         \\
         \bottomrule
    \end{tabular}
    }
    \label{tab:mau_compare_qwen_ap}
\end{table}

\subsection{Robustness to Pixel Perturbations}
\textbf{Robustness under JPEG Compression.}
Tabs.~\ref{tab:fake_jpeg50_qwen}, \ref{tab:fake_jpeg75_qwen}, and \ref{tab:fake_jpeg90_qwen} report robustness results under JPEG compression with quality factors (QF) of 50, 75, and 90, respectively.
Overall, detector performance decreases as compression becomes stronger, especially when QF drops from 90 to 50.
This trend suggests that JPEG compression suppresses high-frequency traces that are commonly exploited by AI-generated image detectors.
Methods such as Forgelens~\cite{chen2025forgelens} and DRCT~\cite{chen2024drct} show particularly large drops, indicating a stronger dependence on compression-sensitive low-level artifacts.
In contrast, FPD consistently maintains strong performance across all compression levels and generation models.
Even under severe compression (QF=50), FPD achieves competitive accuracy on challenging generators such as SD3.5 (73.82\%), FLUX.2 (81.73\%), and Z-Image (84.71\%), while remaining near saturation on Qwen-Image, Nano Banana 2, and GPT-Image 2 when compression is weaker.
Compared with methods that perform well on specific generators but degrade on others, such as B-Free~\cite{bias-free} and LTD~\cite{ltd}, FPD shows more balanced performance across all seven generators.
These results indicate that FPD is less dependent on fragile pixel-level artifacts and better preserves detection ability under realistic image transmission and social-media recompression scenarios.

\begin{table}[htbp]
    \centering
    \vspace{-4mm}
    \caption{Robustness under JPEG compression (QF=50). Average accuracy (\%) on FakeBench using the constructed Qwen-Image-Train dataset for training.}
    \resizebox{0.9\textwidth}{!}{
    \begin{tabular}{lcccccccccccccc}
    \toprule
        \textbf{Methods}& \textbf{SD3.5} &\textbf{FLUX.1} &\textbf{FLUX.2} & \textbf{Z-Image} 
        & \textbf{Qwen-Image} & \textbf{N.B. 2} & \textbf{GPT-I. 2}
        \\
        \midrule
        UniFD &58.36 &72.76   &61.04  &65.68  &81.40 &77.76  &79.56  \\
        LOTA &55.33 &54.80 &53.40 &53.59 &61.36 &54.29 &50.91\\
        DRCT &53.31 &55.55   &56.60  &56.36  &71.99 &54.49  &58.21   \\
        NPR  &54.33 &52.28   &60.73  &60.46  &87.60 &69.80  &89.05  \\
       AIDE &53.71 &57.36   &55.70  &62.24  &95.27 &67.76  &96.35  \\
       Effort &77.49 &82.01   &79.07  &83.03  &98.41 &87.55  &92.15 \\
        Forgelens &52.00 &50.17   &51.12  &51.34  &75.97 &51.22  &52.55 \\
        LTD &69.43 &87.67   &75.40  &76.09   &96.30 &76.73  &66.42   \\
        Aligned &66.93 &61.30   &64.42   &59.01  &98.15 &61.84  &96.90  \\
         B-Free &69.25 &66.33   &81.48  &79.64  &99.99 &100.00  &99.27 \\
         OmniAID  &69.41 &76.83 &71.60 &68.94 &80.37 &72.65 &73.81\\
         DDA &53.10 &53.72  &52.24  &51.81  &75.75 &51.84  &61.13   \\
         DINO-Det &76.83 &92.70   &75.37  &77.99  &96.25 &81.63  &76.82  \\
         \midrule
        \rowcolor{mygray}\textbf{FPD~(Ours)} &73.82 &92.94  &81.73  &84.71   &97.28 &93.07  &97.62   \\
         \bottomrule
    \end{tabular}
    }

    \label{tab:fake_jpeg50_qwen}
\end{table}  

\begin{table}[]
    \centering
    \vspace{-4mm}
    \caption{Robustness under JPEG compression (QF=75). Average accuracy (\%) on FakeBench using the constructed Qwen-Image-Train dataset for training.}
    \resizebox{0.9\textwidth}{!}{
    \begin{tabular}{lcccccccccccccc}
    \toprule
        \textbf{Methods}& \textbf{SD3.5} &\textbf{FLUX.1} &\textbf{FLUX.2} & \textbf{Z-Image} 
        & \textbf{Qwen-Image} & \textbf{N.B. 2} & \textbf{GPT-I. 2}
        \\
        \midrule
        UniFD &65.29 &81.47  &69.33  &75.49  &87.93 &86.33  &85.77  \\
        LOTA &55.40 &54.84 &53.62 &54.00 &61.54 &54.29 &50.91\\
        DRCT &55.49 &62.24  &59.60  &59.85  &71.62 &54.90  &58.21    \\
        NPR  &81.25 &76.33  &86.85  &86.11  &98.88 &95.71  &95.44  \\
       AIDE &63.12 &67.01  &64.42  &72.79  &98.01 &82.24  &95.99   \\
       Effort &83.56 &87.98  &84.63  &89.61  &99.17 &97.14  &95.26  \\
        Forgelens &68.08 &73.18  &58.02  &61.23  &99.28 &70.20  &62.23   \\
        LTD &90.27 &95.50  &90.62  &90.64  &99.43 &96.53  &90.69  \\
        Aligned &85.44 &79.22  &77.78  &62.69  &90.10 &77.55  &88.32   \\
         B-Free &82.74 &83.78  &90.33  &89.65  &99.99 &100.00  &100.00  \\
         OmniAID  &68.89 &76.81 &72.79 &69.83 &81.90 &72.96 &77.00\\
         DDA &74.53 &81.07  &67.68  &66.45  &97.29 &79.59  &78.83   \\
         DINO-Det &88.11 &95.95  &85.72  &88.21  &98.40 &93.27  &85.77    \\
         \midrule
        \rowcolor{mygray}\textbf{FPD~(Ours)} &86.47 &96.81  &92.42  &93.16  &99.75 &99.22  &98.91  \\
         \bottomrule
    \end{tabular}
    }

    \label{tab:fake_jpeg75_qwen}
\end{table}  

\begin{table}[htbp]
    \centering
    \vspace{-4mm}
    \caption{Robustness under JPEG compression (QF=90). Average accuracy (\%) on FakeBench using the constructed Qwen-Image-Train dataset for training.}
    \resizebox{0.9\textwidth}{!}{
    \begin{tabular}{lcccccccccccccc}
    \toprule
        \textbf{Methods}& \textbf{SD3.5} &\textbf{FLUX.1} &\textbf{FLUX.2} & \textbf{Z-Image} 
        & \textbf{Qwen-Image} & \textbf{N.B. 2} & \textbf{GPT-I. 2}
        \\
        \midrule
        UniFD &65.28 &80.90  &68.59  &74.50  &86.16 &85.51  &86.68    \\
        LOTA &55.33 &55.02 &53.32 &54.03 &61.77 &55.71 &52.19\\
        DRCT &56.17 &62.78  &60.74  &60.51  &72.75 &54.29  &57.48   \\
        NPR  &80.91 &75.44  &86.61  &85.73  &98.81 &95.92  &97.08   \\
       AIDE &63.54 &67.42  &64.61  &73.64  &97.60 &81.22  &96.17   \\
       Effort &84.12 &88.42  &85.24  &90.25  &99.10 &97.14  &96.35   \\
        Forgelens &85.61 &84.64  &76.16  &75.77  &94.83 &91.02  &78.10   \\
        LTD &91.31 &94.23  &93.01  &92.78  &95.97 &94.69  &90.15    \\
        Aligned &83.54 &77.53 &76.92 &73.36 &87.57 &75.71 &87.77  \\
         B-Free &81.68 &82.42  &89.75  &89.02  &99.99 &100.00  &100.00  \\
         OmniAID &67.43 &75.09 &71.30 &68.40 &79.41 &71.54 &74.27\\
         DDA &80.30 &85.06  &73.57  &71.34  &95.78 &86.12  &95.07   \\
         DINO-Det &90.33 &97.13  &88.31  &90.40  &98.02 &93.47  &94.89   \\
         \midrule
        \rowcolor{mygray}\textbf{FPD~(Ours)} &90.40 &96.92  &93.21  &93.11  &99.83 &98.77  &99.27   \\
         \bottomrule
    \end{tabular}
    }
    \vspace{-4mm}
    \label{tab:fake_jpeg90_qwen}
\end{table}  

\begin{table}[htbp]
    \centering
    \vspace{-4mm}
    \caption{Robustness under Gaussian blur ($\sigma=1$). Average accuracy (\%) on FakeBench using the constructed Qwen-Image-Train dataset for training.}
    \resizebox{0.9\textwidth}{!}{
    \begin{tabular}{lcccccccccccccc}
    \toprule
        \textbf{Methods}& \textbf{SD3.5} &\textbf{FLUX.1} &\textbf{FLUX.2} & \textbf{Z-Image} 
        & \textbf{Qwen-Image} & \textbf{N.B. 2} & \textbf{GPT-I. 2}
        \\
        \midrule
         UniFD &66.88   &82.94   &76.59   &71.17  &88.49  &87.14   &88.50   \\
         LOTA &53.54 &56.31 &53.92 &52.88 &57.96 &55.51 &55.11\\
        DRCT &56.83  &61.29   &57.65   &57.74  &69.26  &57.55   &55.47   \\
        NPR  &83.72  &79.50   &87.36   &88.31  &98.27  &94.49   &96.90    \\
       AIDE &74.19  &85.52   &83.01   &80.10  &90.82  &91.63   &87.41     \\
       Effort &87.08  &92.74   &92.31    &87.04  &98.69   &97.35   &96.35     \\
        Forgelens &66.31  &69.25   &62.32   &65.49  &71.14  &69.39   &70.80    \\
        LTD &92.06  &95.51   &91.83   &91.90  &94.95  &93.27   &86.86  \\
        Aligned &86.09  &82.68    &73.01   &70.46  &95.79  &84.29   &96.17      \\
         B-Free &82.66  &82.44   &83.62   &84.12  &86.76  &82.86   &86.31    \\
         OmniAID  &68.26 &74.92 &71.27 &68.79 &78.25 &69.59 &67.63\\
         DDA &81.46  &86.23   &74.98   &76.09   &94.33  &89.39   &94.71      \\
         DINO-Det &90.51  &96.85   &91.10   &89.14  &97.54  &93.27   &95.44    \\
         \midrule
        \rowcolor{mygray}\textbf{FPD~(Ours)} &92.50  &98.48   &94.89   &94.14  &99.44   &98.57   &98.91     \\
         \bottomrule
    \end{tabular}
    }

    \vspace{-4mm}
    \label{tab:fake_blur1_qwen}
\end{table}  

\begin{table}[htbp]
    \centering
    \vspace{-4mm}
    \caption{Robustness under Gaussian blur ($\sigma=3$). Average accuracy (\%) on FakeBench using the constructed Qwen-Image-Train dataset for training.}
    \resizebox{0.9\textwidth}{!}{
    \begin{tabular}{lcccccccccccccc}
    \toprule
        \textbf{Methods}& \textbf{SD3.5} &\textbf{FLUX.1} &\textbf{FLUX.2} & \textbf{Z-Image} 
        & \textbf{Qwen-Image} & \textbf{N.B. 2} & \textbf{GPT-I. 2}
        \\
        \midrule
         UniFD &66.90  &83.38   &76.69   &72.25  &89.58  &87.76   &89.23   \\
         LOTA &53.02 &55.79 &53.32 &52.10 &56.61 &51.84 &59.85\\
        DRCT &56.68  &60.87   &57.68   &57.95  &69.25  &57.35   &56.02    \\
        NPR  &83.95  &80.82   &86.01   &87.16  &94.64  &90.20   &93.98     \\
       AIDE &68.80  &80.67   &80.68   &74.82  &92.72  &86.53   &88.87    \\
       Effort &89.77  &94.35   &93.14   &87.96  &96.77  &94.49   &93.43   \\
        Forgelens &52.26  &52.44    &51.14   &52.45  &52.63  &51.63   &52.55     \\
        LTD &79.34  &81.16   &78.59  &79.69  &80.46  &77.35   &75.36    \\
        Aligned &76.75  &82.37   &74.18   &75.21  &89.06  &85.71    &91.06   \\
         B-Free &66.81  &66.10   &66.47   &68.04  &68.70   &64.69   &69.71    \\
         OmniAID &68.50 &74.37 &70.86 &69.57 &77.63 &68.16 &66.36\\
         DDA &79.80  &85.24   &73.53   &74.94  &89.05  &87.55   &88.87   \\
         DINO-Det &89.04  &95.28   &89.81   &88.50  &95.33  &91.02   &93.07\\
         \midrule
        \rowcolor{mygray}\textbf{FPD~(Ours)} &87.53  &97.24   &93.17   &91.60  &99.18  &97.75   &99.08   \\
         \bottomrule
    \end{tabular}
    }

    \label{tab:fake_blur3_qwen}
\end{table}  

\textbf{Robustness under Gaussian Blur.}
Tabs.~\ref{tab:fake_blur1_qwen}, \ref{tab:fake_blur3_qwen}, and \ref{tab:fake_blur5_qwen} report detector performance under Gaussian blur with $\sigma=1$, $\sigma=3$, and $\sigma=5$, respectively.
Gaussian blur substantially affects detectors that rely on local high-frequency cues.
As blur strength increases, several methods suffer severe degradation; for example, when $\sigma \geq 3$, LTD~\cite{ltd} and B-Free~\cite{bias-free} show clear performance declines.
This behavior suggests that these detectors depend heavily on texture-level artifacts that are easily weakened by image smoothing.
In contrast, FPD remains stable across different blur levels.
From $\sigma=1$ to $\sigma=5$, FPD shows only limited degradation for most generation models, consistently achieving over 91\% accuracy on SD3.5, FLUX.1, FLUX.2, and Z-Image, and maintaining near-saturated performance on Qwen-Image, Nano Banana 2, and GPT-Image 2.
Moreover, the performance gap between FPD and existing methods becomes larger under stronger blur.
At $\sigma=5$, several competitive detectors experience accuracy reductions exceeding 10\%, whereas FPD remains close to its performance under mild blur.
These results demonstrate that FPD is more robust to post-processing-induced distribution shifts and does not rely solely on superficial high-frequency generation artifacts.

\begin{table}[htbp]
    \centering
    \vspace{-4mm}
    \caption{Robustness under Gaussian blur ($\sigma=5$). Average accuracy (\%) on FakeBench using the constructed Qwen-Image-Train dataset for training.}
    \resizebox{0.9\textwidth}{!}{
    \begin{tabular}{lcccccccccccccc}
    \toprule
        \textbf{Methods}& \textbf{SD3.5} &\textbf{FLUX.1} &\textbf{FLUX.2} & \textbf{Z-Image} 
        & \textbf{Qwen-Image} & \textbf{N.B. 2} & \textbf{GPT-I. 2}
        \\
        \midrule
        UniFD &66.79  &83.36   &76.65   &72.26  &89.69  &87.76   &89.42  \\
        LOTA &52.77 &55.45 &53.23 &51.98 &56.54 &51.43 &60.58\\
        DRCT &56.66  &61.12   &57.92   &57.70  &69.22  &58.78   &55.29    \\
        NPR  &83.65  &80.64   &85.51   &86.49  &93.68  &89.59   &92.88    \\
       AIDE &66.62  &77.90   &78.53   &71.88  &93.15  &84.49   &89.05   \\
       Effort &89.76  &94.29   &93.00   &87.97  &96.41  &94.08   &93.61    \\
        Forgelens &51.20  &51.36   &50.57   &51.28  &51.44  &51.63   &51.46   \\
        LTD &78.59  &80.53   &77.86   &79.03  &79.80  &75.10   &73.18    \\
        Aligned &75.78  &81.93   &73.55   &75.07  &88.21  &83.67   &90.15    \\
         B-Free &70.18  &69.46   &69.75   &71.12  &71.81  &67.35   &71.72 \\
         OmniAID &68.23 &74.04 &70.32 &69.44 &77.32 &66.73 &66.18\\
         DDA &78.91  &83.58   &72.93   &74.47  &86.37  &84.29   &87.04  \\
         DINO-Det &88.83  &94.74   &89.56   &88.17  &94.70  &90.00   &92.34    \\
         \midrule
        \rowcolor{mygray}\textbf{FPD~(Ours)} &87.11  &97.36   &93.34   &91.05  &99.16  &97.55   &97.99    \\
         \bottomrule
    \end{tabular}
    }

    \label{tab:fake_blur5_qwen}
\end{table}

\section{Related Work}

\subsection{AI-Generated Image Detection Benchmarks}

Benchmarks for AI-generated image detection (AIID) have evolved alongside image generation models.
Early datasets, such as CNNSpot~\cite{wang2020cnn}, mainly collect fake images produced by GAN-based generators~\cite{karras2019style,brock2018large}.
With the rapid progress of diffusion models~\cite{ho2020denoising,sd}, synthetic images have become increasingly realistic, making real/fake discrimination more challenging.
To support evaluation under diffusion-based generation, several benchmarks have been proposed, including GenImage~\cite{genimage}, AIGCDetect~\cite{AIGCDetect}, DiffusionDB~\cite{wang2023diffusiondb}, and ArtiFact~\cite{rahman2023artifact}.
For example, GenImage covers ImageNet's 1,000 classes and includes images generated by eight representative models from both academia, such as Stable Diffusion~\cite{sd}, and industry, such as Midjourney~\cite{Midjourney}.

Recent benchmarks further move toward in-the-wild and more realistic evaluation settings.
Chameleon~\cite{chameleon} and WildRF~\cite{wildrf} collect AI-generated content from online platforms, introducing distribution shifts caused by real-world image sharing.
RRDataset~\cite{rrdataset} evaluates detectors across multiple real-world scenarios using recent generators such as SD3.5 Large~\cite{sd3.5} and FLUX.1~\cite{flux2024}.
AIGIBench~\cite{aigibench} expands benchmark coverage by including diverse data sources, including GAN-based images~\cite{karras2019style,chen2020simswap,zhang2022styleswin}, diffusion-generated images~\cite{podell2024sdxl,flux2024}, personalized generation models~\cite{wang2024instantid,ye2023ip}, and images from online platforms such as X, Facebook, and Reddit.
Despite these advances, existing benchmarks still primarily focus on fully synthetic images and provide limited coverage of localized image manipulation.
Moreover, many benchmarks rely on outdated generators or restricted real-image sources, which limits their ability to reflect the distribution of modern generated and edited images.
In contrast, DailyBench is designed to evaluate both full-image synthesis and object-level manipulation using modern open-source and commercial generative models.

\subsection{AI-Generated Image Detection}

AI-generated image detection methods can be broadly grouped into representation-based detectors, artifact-based detectors, and robustness-oriented training strategies.
Early methods, such as CNNSpot~\cite{wang2020cnn}, use vanilla CNN classifiers and show that detectors can perform well on seen generators but often fail to generalize to unseen ones.
To improve cross-generator generalization, UniFD~\cite{unifd} adopts a CLIP backbone, while subsequent methods~\cite{tan2024frequency,effort,zheng2024breaking} explore stronger architectures and preprocessing strategies.
C2P-CLIP~\cite{tan2025c2p} further adapts CLIP by injecting explicit ``real'' and ``fake'' concepts for AIGI detection.
These methods improve semantic representation learning, but their robustness under modern generation and manipulation settings remains insufficiently evaluated.

Another line of work detects synthetic images by exploiting artifact cues in the spatial or frequency domain.
Frequency-based methods show that subtle high-frequency patterns are useful for separating real and synthetic images~\cite{chen2024drct,safe,karageorgiou2025any}.
NPR~\cite{npr} focuses on upsampling artifacts, while DDA~\cite{dda} mixes real and synthetic images in both pixel and frequency domains to create more realistic training samples.
More recently, DINO-Det~\cite{dinodet} improves robustness under image degradation by distilling feature and logit responses from sharp images to blurred images using a frozen teacher model.
ForgeLens~\cite{chen2025forgelens} guides the extraction of forgery-specific features during training and effectively integrates forgery information from multi-stage features to enable a frozen network to focus on forgery-specific features.
LTD~\cite{ltd} captures the inter-layer consistency differences between real and fake images, adaptively identifies the most discriminative layers, and evaluates the transition differences between layers to construct discriminative features.
PROBE~\cite{probe} uses detector feedback to guide a generator toward realistic, hard-to-classify samples and then incorporates these samples into detector training.
Collectively, prior methods have advanced cross-generator detection and robustness. 
However, their evaluations remain centered largely on fully synthesized images and provide limited evidence on high-fidelity outputs from recent generators or localized object-level edits. 
DailyBench complements these efforts by evaluating detectors across both modern full-image synthesis and localized manipulation settings.

\subsection{AI-Manipulation Image Localization}
Image manipulation localization identifies altered regions, often alongside image-level authenticity prediction. 
UnionFormer~\cite{li2024unionformer} combines RGB, noise, and object-consistency cues for joint detection and localization. 
To generalize to unseen manipulations, GIFL~\cite{zhao2025learning} learns features of pristine content rather than forgery-specific traces, while IFA-Net~\cite{zhang2026detecting} uses reconstruction deviations from a real-image prior to refine localization. 
RITA~\cite{zhu2026revisiting} addresses multi-step manipulation by predicting edited regions sequentially instead of producing a single localization mask.

Recent benchmarks broaden the editing scenarios considered in localization research. 
GIM~\cite{chen2025gim} provides a large-scale benchmark of generative manipulations and introduces GIMFormer, while OpenSDI~\cite{wang2025opensdi} introduces OpenSDID and MaskCLIP for localizing diffusion-generated manipulations.
DEAL-300K~\cite{zhang2025deal} covers mask-free diffusion edits, DiffSeg30k~\cite{ci2025diffseg30k} covers multi-turn editing, and LocateEdit-Bench~\cite{wu2026locateedit} evaluates instruction-based edits. 
%
Their localization analyses complement DailyBench's focus on image-level detector generalization across full-image synthesis and localized manipulation.


\end{document}